%% file: main.tex
\documentclass[graybox]{svmult}

\usepackage{mathptmx}       % selects Times Roman as basic font
\usepackage{helvet}         % selects Helvetica as sans-serif font
\usepackage{courier}        % selects Courier as typewriter font
\usepackage{type1cm}        % activate if the above 3 fonts are
\usepackage{makeidx}         % allows index generation
\usepackage{graphicx}        % standard LaTeX graphics tool
\usepackage{multicol}        % used for the two-column index
\usepackage[bottom]{footmisc}% places footnotes at page bottom
\usepackage{natbib}
\usepackage[]{algorithm2e}
\usepackage{hyperref}

\makeindex             % used for the subject index
\begin{document}

\title*{An Exploration of Exploration: Measuring the ability of lexicase selection to find obscure pathways to optimality}
%%%%%%%%%%%%%%%%%%%%%%%%%%%%%
% Title ideas
%  Diagnosing fitness landscape exploration in/by lexicase selection
%  how different variants of lexicase selection affect exploration
%  Diagnosing lexicase selection's fitness landscape exploration
% Measuring the Exploration Capacity of Lexicase Selection and its Variants
% ... with diagnostic
% Diagnosing the Exploration Capacity of Lexicase Selection and its Variants
% Using Diagnostics to Measure the Exploration Capacity of Lexicase Selection and its Variants
% A Diagnostic for Measuring the Exploration Capacity of Lexicase Selection and its Variants
% Exploring exploration capacity of lexicase and its variants
% Exploring the capacity for exploration  of lexicase and its variants
% An Exploration of Exploration: Measuring the ability of lexicase selection to find obscure pathways to optimality
% Diagnostic to Measure Exploration Capacity of Lexicase and its variants
%%%%%%%%%%%%%%%%%%%%%%%%%%%%%

% Shortened title (if full title is too long)
\titlerunning{An exploration of exploration} 

\author{
Jose Guadalupe Hernandez, 
Alexander Lalejini, 
and Charles Ofria
}

\authorrunning{Hernandez, Lalejini, and Ofria}

\institute{Jose Guadalupe Hernandez \at Michigan State University, East Lansing, MI, USA \\ \email{herna383@msu.edu}
\and Alexander Lalejini \at Michigan State University, East Lansing, MI, USA \\ \email{amlalejini@gmail.com}
\and Charles Ofria \at Michigan State University, East Lansing, MI, USA \\ \email{ofria@msu.edu}}
\maketitle

\abstract*{\input{tex/abstract}}

\abstract{\input{tex/abstract}}

\input{tex/introduction}

\input{tex/exploration-diagnostic}

\input{tex/lexicase-selection}

\input{tex/diagnosing-lexicase-and-its-variants}

\input{tex/conclusion}

\input{tex/software-data-availability}

\input{tex/acknowledgements}

\bibliographystyle{apalike}
\bibliography{references,software}

\end{document}

%% file: tex/abstract.tex
Parent selection algorithms (selection schemes) steer populations through a problem's search space, often trading off between exploitation and exploration. %, which, in turn, influences problem-solving success.
Understanding how selection schemes affect exploitation and exploration within a search space is crucial to tackling increasingly challenging problems.
Here, we introduce an ``exploration diagnostic'' that diagnoses a selection scheme's capacity for search space exploration.
We use our exploration diagnostic to investigate the exploratory capacity of lexicase selection and several of its variants: epsilon lexicase, down-sampled lexicase, cohort lexicase, and novelty-lexicase.
% @AML: Here's a list of the big takeaways in sentence form. NEEDS STREAMLINING/TRIMMING!
We verify that lexicase selection out-explores tournament selection, and we show that lexicase selection's exploratory capacity can be sensitive to the ratio between population size and the number of test cases used for evaluating candidate solutions.
Additionally, we find that relaxing lexicase's elitism with epsilon lexicase can further improve exploration. 
Both down-sampling and cohort lexicase---two techniques for applying random subsampling to test cases---degrade lexicase's exploratory capacity; however, we find that cohort partitioning better preserves lexicase's exploratory capacity than down-sampling.
Finally, we find evidence that novelty-lexicase's addition of novelty test cases can degrade lexicase's capacity for exploration.
Overall, our findings provide hypotheses for further exploration and actionable insights and recommendations for using lexicase selection.
Additionally, this work demonstrates the value of selection scheme diagnostics as a complement to more conventional benchmarking approaches to selection scheme analysis.

%% file: tex/introduction.tex
\section{Introduction}
\label{sec:introduction}

% -- Lexicase selection very good at exploration --
Lexicase-based parent selection algorithms have proven to be highly successful for finding effective solutions to test-based problems in genetic programming (GP)~\citep{helmuth_general_2015,orzechowski_where_2018,helmuth2020benchmarking}.
Lexicase selection's success is rooted in its ability to balance strong search space exploration with simultaneous exploitation.  
That is, lexicase selection maintains meaningfully diverse populations~\citep{helmuth_effects_2016,helmuth_importance_2020} by promoting the coexistence of subpopulations that are each focused on different aspects of a problem (\textit{e.g.}, on different test cases or selection criteria)~\citep{dolson_ecological_2018}.
As such, lexicase selection algorithms are able to explore many promising problem-solving pathways in parallel, optimizing each until an overall solution is found. 

% -- Types of problems where lexicase is useful + how it works --
Many genetic programming problems are multi-faceted where the quality of a candidate solution must be measured according to its performance on a set of test cases.
For such problems, we must decide how to combine performances across many test cases in order to select promising individuals to produce offspring for the next generation.  
Traditional parent selection algorithms assess the quality of an individual by aggregating their performance on all test cases. 
The lexicase selection algorithm, however, chooses each parent based on the relative performances of candidate solutions on random permutations of the test set. 
Specifically, each time a parent is needed, the entire population is considered as candidates for selection, and the full set of test cases are shuffled; each test case is applied sequentially (in the given shuffled order) to the current set of candidates, removing all but the best candidates from consideration until only a single individual remains to be selected~\citep{helmuth2015solving}.
Because the ordering of test cases is different for each parent selection event, individuals that perform well on different subsets of problems are able to coexist~\citep{dolson_ecological_2018}. 
Moreover, lexicase selection exerts strong selection pressure to optimize each subpopulation, as only the best candidates on different sequences of test cases are selected.

% -- Variants of lexicase selection --
Indeed, the successes of the original lexicase selection algorithm have inspired numerous variants, each either specialized for solving different categories of problems or designed to address potential shortcomings of the original lexicase algorithm (\textit{e.g.}, computational efficiency).
Such variants include epsilon lexicase \citep{la2016epsilon,la_cava_probabilistic_2019}, down-sampled lexicase~\citep{hernandez2019random}, novelty-lexicase~\citep{jundt_comparing_2019}, ALPS lexicase~\citep{helmuth2020benchmarking}, and batch-lexicase selection~\citep{aenugu_lexicase_2019}.
Many of these variants have been rigorously benchmarked on their problem-solving success and on their ability to maintain phenotypic and phylogenetic diversity~\citep{ferguson2020characterizing,helmuth_lexicase_2016,helmuth_effects_2016,la_cava_relaxations_2018}.
However, benchmarking is often performed in the context of a particular GP system and with the overall goal of measuring performance on challenging computational problems (\textit{e.g.}, program synthesis benchmark problems from~\citealt{helmuth_general_2015} and~\citealt{helmuth_psb2_2021}). 
While such benchmarking is critical for understanding the real-world applicability of a selection scheme, the specific problems used do not always allow us to disentangle the particular pros and cons of each scheme~\citep{hooker_testing_1995}.
For this paper, we focus on one important aspect of lexicase-based selection schemes: How do we isolate the \textit{exploration} capabilities of lexicase selection and its variants?

% -- Our aims --
We introduce an ``exploration diagnostic'' and use it to test how well a set of parent selection algorithms can explore a simple landscape with many uphill pathways of differing peak fitnesses.
Our exploration diagnostic allows for the total number of possible evolutionary pathways to be tuned, enabling practitioners to find where an algorithm's exploratory abilities begin to fall off.
First, we verify established expectations that lexicase selection better facilitates search space exploration than tournament selection, a more traditional selection algorithm. 
Next, we evaluate lexicase selection on our exploratory diagnostic with an increasing number of possible pathways identify its exploratory limitations.
Finally, we apply our exploration diagnostic to four variants of lexicase selection: epsilon lexicase, down-sampled lexicase, cohort lexicase, and novelty-lexicase selection. 

% -- Results --
We find that lexicase selection drives performance improvement at each of the exploration diagnostic difficulty levels that we evaluated.
Lexicase selection finds nearly perfect solutions for fitness landscapes with a small number of pathways to be explored, and performance gradually declines as the number of possible evolutionary pathways increases.
Additionally, we show that lexicase selection can be sensitive to the ratio between population size and the number of test cases used for evaluating candidate solutions.
For small values of $\epsilon$, epsilon lexicase improves the exploratory capacity of lexicase selection.
Random subsampling via either down-sampled or cohort lexicase degrades exploratory capacity, but cohort partitioning better preserves lexicase's exploratory capacity than down-sampling.
Finally, we did not find compelling evidence that novelty-lexicase improves performance on the exploration diagnostic relative to standard lexicase selection; in fact, the addition of novelty test cases can substantially degrade lexicase's diagnostic performance.

%% file: tex/exploration-diagnostic.tex
\section{Exploration diagnostic}
\label{sec:exploration-diagnostic}
% Diagnosing the capacity for exploration

% Why is exploration important
Understanding how parent-selection algorithms affect exploration and exploitation within a search space is crucial to tackling increasingly challenging problems.
This information can help determine what modifications to an evolutionary algorithm may be needed to improve the likelihood of finding a high quality solution.
Different selection schemes (or other components of an evolutionary algorithm) can alter the trade-off between exploitation and exploration \citep{eiben1998evolutionary}.
An exploitation-only selection scheme will push the population to the closest optimum and not allow it to explore other promising regions of the search space.
Conversely, an exploration-only selection scheme will scatter the population across the entire search space but is unlikely to reach nearby optima. 
Hence, striking a balance between exploration and exploitation is critical to finding high-quality solutions. 
Here, we introduce a diagnostic that challenges selection schemes to explore multiple avenues of a search space, each with an upward pathway, with the goal of finding the best avenue to hill climb.

\input{tex/figures/fig-genotype-phenotype}

% How does the diagnostic work
We balanced both exploitation and exploration in our diagnostic.
Specifically, we designed a problem with many upward pathways that all have identical slopes, but vary in total length.
Since shorter pathways are always equivalent to the beginning of longer pathways, exploration is critical for finding the longest pathway (which will lead to the global optimum).
In the end, the only way for an evolving population to determine the length of a pathway is to follow it.

Candidate solutions for this diagnostic are numerical vectors of a designated size (its ``cardinality'' -- we used 100 as the default cardinality in this work).
Cardinality determines the number of pathways to local optima in the fitness landscape.
Each value in a candidate solution is a floating-point number between 0.0 and 100.0.
To evaluate a candidate solution, we first scan its vector to find the maximum value and designate its position as the ``activation position'' for calculating its fitness.
From an intuitive perspective, the activation position defines which peak the candidate solution is climbing toward. 
Beginning at the activation position, we sum all consecutive values that are less than or equal to each previous position.  
We stop when either a position is no longer monotonically non-increasing or we reach the end of the vector.
We refer to this consecutive sequence of scored values as the ``active region'' of the candidate solution.
All values outside of the active region have zero fitness contribution.
The fitness contributions of each position (\textit{i.e.}, each trait) define the ``phenotype'' of the candidate solution; two candidate solutions that differ only in inactive regions will have identical phenotypes.
Figure \ref{fig:genotype-phenotype} shows an example fitness calculation.
Given this search space, the optimal solution will have a 100.0 in every position of its vector starting from the very first, making the entire candidate solution active and each value maximized.
However, any candidate solution with an activation position other than the first will not have a pathway to the global optimum that is reachable via hill climbing alone.

% Any extra details 
Given the large number of pathways that need to be simultaneously explored, this diagnostic allows us to compare the exploration capacity of different selection schemes.
Additionally, this diagnostic allows researchers to test the exploration breaking point of a given selection scheme, as increasing the cardinality of the diagnostic increases the exploratory capacity needed to find the best activation position.
In this work, we use this diagnostic to test the exploratory limits of lexicase selection along with a number of its variants.

%% file: tex/figures/fig-genotype-phenotype.tex
\begin{figure}[ht!]
\centering
\includegraphics[width=\textwidth]{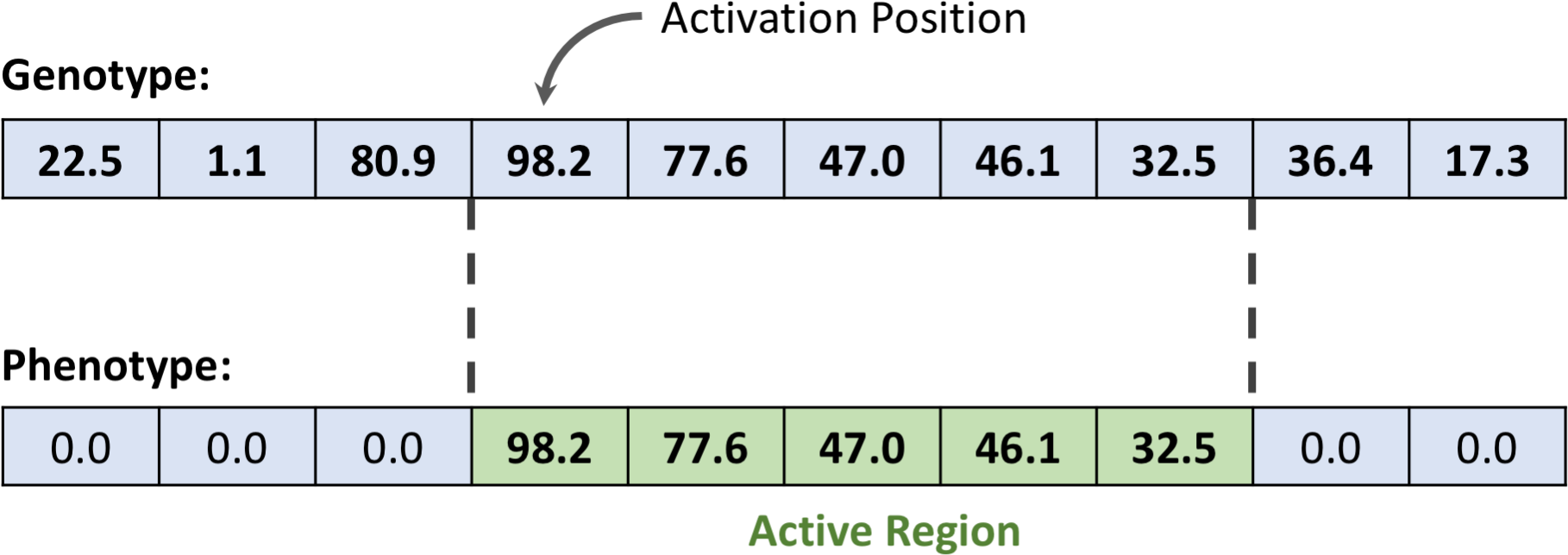}
\caption{\textbf{An example evaluation with the exploration diagnostic.}  
A candidate solution with a cardinality of 10 is analyzed.  
The highest value in its vector is identified as 98.2, and its position is marked as the beginning of the active region.  
The next four values are all in a decreasing sequence (77.6, 47.0, 46.1, and 32.5) and are thus all considered part of the active region.  
The value after that (36.4) is greater than its predecessor and thus left inactive, closing the active region.  
All values not in the active region are expressed in the phenotype as 0.0.  
The total fitness of the sequence is the sum of the values in the phenotype or 0.0 + 0.0 + 0.0 + 98.2 + 77.6 + 47.0 + 46.1 + 32.5 + 0.0 + 0.0 = 301.4.}
\label{fig:genotype-phenotype}
\end{figure}

%% file: tex/lexicase-selection.tex
\section{Lexicase selection}
\label{sec:lexicase-selection}

\input{tex/figures/alg-lexicase}

\cite{spector_assessment_2012} introduced the lexicase parent selection algorithm for solving GP problems that require programs to produce qualitatively different modes of response for different inputs.
Since its introduction, lexicase selection has been demonstrated to be successful across a broad range of problem domains, including automatic program synthesis~\citep{helmuth_general_2015}, 
symbolic regression~\citep{la2016epsilon}, 
evolutionary robotics~\citep{moore_lexicase_2017}, 
genetic algorithms~\citep{metevier_lexicase_2019}, 
and learning classifier systems~\citep{aenugu_lexicase_2019}.

In lexicase selection, individuals are evaluated on a set of selection criteria (\textit{e.g.}, test cases or other types of fitness functions).
For each selection event, each member of the population is initially considered to be a candidate for selection.
To select an individual, lexicase shuffles the set of test cases, and considers each test case in sequence. 
In shuffled order, each test case is used to filter the candidates, removing all but the best individuals from further consideration.
This process of winnowing candidates continues until only one candidate remains to be selected or until all test cases have been considered; if more than one candidate remains, one is selected at random.
Algorithm~\ref{alg:lexicase} details the lexicase selection algorithm.

Because the set of test cases are shuffled whenever a parent must be chosen, individuals that perform well on different partitions of the test set can coexist within the population~\citep{dolson_ecological_2018}.
Indeed, this dynamic creates niches where different members of the population can specialize on different subsets of selection criteria, allowing a population to simultaneously explore many pathways to solving a given problem. 
Moreover, this focus on exploration does not necessarily sacrifice lexicase's ability to exploit each pathway since only the best performing individuals are selected for a given sequence of test cases. 

Many variants of lexicase selection have been proposed, each either specialized for solving a particular type of problem or designed to address potential short comings of the original lexicase selection scheme. 
Below, we describe each of the four variants of lexicase selection examined in this work. 

\subsection{Epsilon lexicase selection}
\label{sec:epsilon-lexicase-selection}

Epsilon lexicase selection relaxes the elitism of the filtering step in standard lexicase selection (step 3c in Algorithm~\ref{alg:lexicase}).
When filtering candidates on a given test case, epsilon lexicase retains all individuals with performances within some threshold ($\epsilon$) of the best performance on that test case.
The $\epsilon$ parameter can be tuned by the practitioner and can be applied either as a proportion of the optimal performance on a given test case or as an absolute threshold.
% In this work, we apply $\epsilon$ as an ab; however, more dynamic ways of configuring 

Epsilon lexicase selection specializes standard lexicase selection for problems where performances on selection criteria are measured using real-valued numbers, such as symbolic regression problems \citep{la2016epsilon,la_cava_relaxations_2018,orzechowski_where_2018}
or evolving robot controllers~\citep{moore_comparison_2016,moore_lexicase_2017}.
The standard lexicase selection algorithm assumes that individuals with equivalent performances on a given test case will have equal scores for that test case.
Inconsequential noise in an individual's score on a particular test case could result in arbitrary, but consequential differences in which individuals are selected by the standard lexicase algorithm. 
By allowing a small $\epsilon$ difference between individuals, epsilon lexicase addresses this potential problem. 
% Indeed, epsilon lexicase has been shown to be highly effective for symbolic regression problems \citep{la2016epsilon,}.

In this work, we vary $\epsilon$ to investigate how it affects exploration.
\cite{la2016epsilon} observed that behavioral diversity increases at larger values of $\epsilon$.  
Given $\epsilon$'s affect on behavioral diversity, we hypothesize that increasing $\epsilon$ will increase the exploration capacity of epsilon lexicase.
However, at too high of an $\epsilon$ value, we expect \textit{meaningful} exploration to degrade. 
That is, as $\epsilon$ increases beyond a certain point, different adaptive pathways blur together as meaningful differences in test case performances become indistinguishable.
 
For simplicity, we apply $\epsilon$ as a fixed absolute error threshold in this work.
Future work, however, should investigate how different applications of $\epsilon$ further influence lexicase's exploration capacity (\textit{e.g.}, semi-dynamic and dynamic applications of $\epsilon$ from~\citealt{la_cava_probabilistic_2019}).

\subsection{Down-sampled lexicase selection}
\label{sec:down-sampled-lexicase-selection}

Down-sampled lexicase applies random subsampling to the selection criteria in order to reduce the per-generation computational effort required by lexicase selection~\citep{hernandez2019random,ferguson2020characterizing}. 
Down-sampled lexicase uses a random subset of test cases each generation, which reduces the number of test cases on which each individual in the population must be evaluated every generation.
After down sampling, the standard lexicase procedure is used to choose parents. 

For an equivalent number of total evaluations, down-sampled lexicase allows practitioners to run their evolutionary computing system for more generations or with a larger population size; both of which have been shown to improve problem-solving success~\citep{hernandez2019random,ferguson2020characterizing,helmuth_explaining_2020}.
In this work, we investigate how down sampling affects lexicase selection's exploratory capacity.
While \cite{ferguson2020characterizing} found no evidence that down sampling reduces phenotypic diversity across a range of program synthesis problems, they did find that down sampling degrades specialist maintenance.
We hypothesize that down sampling's negative effect on specialist maintenance harms its exploratory capacity.
Entire categories of test cases may be excluded on any given generation, and candidate solutions specializing on those test cases may be lost as a result. 
Such dynamics may prevent extensive exploration of valuable niches.

\subsection{Cohort lexicase selection}
\label{sec:cohort-lexicase-selection}

Cohort lexicase partitions the test case set and the population each into an equal number of cohorts.
Each generation, cohort membership is randomly assigned, and each cohort of candidate solutions is paired with a cohort of test cases.
Each cohort of candidate solutions is evaluated only on the test cases in the paired test case cohort, which, like down-sampled lexicase, reduces the required number of per-generation evaluations relative to standard lexicase selection.
Unlike down-sampled lexicase, however, cohort lexicase ensures that every test case in the full set is used every generation, as each cohort of candidate solutions competes on a different subset of the full set.
To select a parent, cohort lexicase first selects a cohort to choose from; previous work guaranteed an equal number of parents were selected from each cohort each generation~\citep{hernandez2019random,ferguson2020characterizing}. 
Candidate solutions only compete against other solutions within their respective cohort, and within-cohort competition is arbitrated by the test cases in the associated cohort of tests.

In this work, we investigate how the number of cohorts that we partition the population and test set into influences lexicase selection's capacity for exploration. 
For similar reasons to down-sampled lexicase, we expect cohort lexicase selection to degrade lexicase selection's exploratory capacity.
However, because cohort lexicase uses every test case in every generation, we expect it to better support exploration than down-sampled lexicase.
As we increase the size of cohorts (and decrease the number of cohorts), we expect cohort lexicase to approach the exploratory abilities of standard lexicase selection.
This could be due to the fact that as cohort size increases, the chances of a specialist being paired with the test cases it specializes on also increases. 

\subsection{Novelty-lexicase selection}
\label{sec:novelty-lexicase-selection}

Novelty-lexicase selection combines standard lexicase selection with novelty search \citep{jundt_comparing_2019}. Novelty search disregards functional objectives and instead searches for behavioral novelty, steering populations to continuously explore new regions of the search space~\citep{lehman_abandoning_2011}. 
As such, novelty search is argued to be well-suited for solving problems with deceptive fitness landscapes where local gradients lead \textit{away} from the global optimum~\citep{lehman_exploiting_2008}. 
Novelty-lexicase selection incorporates ideas from novelty search into lexicase selection.

Novelty-lexicase selection (as introduced in~\citealt{jundt_comparing_2019}) requires that the entire population be evaluated on all test cases.
For each member of the population, novelty-lexicase selection computes their ``novelty score'' on each test case.
A novelty score measures how different a candidate solution's output on a given test case is from the rest of the population. 
Here, a candidate solution's novelty score on a test case equals the average distance between its output and the $k$ nearest neighbor outputs for that test case. 
Novelty-lexicase selection incorporates novelty scores by augmenting the test case set with an additional novelty test case for every original test case. 
Using this augmented set of test cases, the standard lexicase procedure is used to choose parents.

In this work, we use our exploration diagnostic to compare the exploratory capacity of novelty-lexicase selection (at $k=$1, 2, 4, 8, 15, 30, and 60) and standard lexicase selection ($k=0$).
\cite{jundt_comparing_2019} found that novelty-lexicase selection generally maintained more behavioral diversity than standard lexicase selection on several program synthesis problems.
As such, we expect the addition of novelty score test cases to improve lexicase selection's exploratory capacity on our exploration diagnostic.

%% file: tex/figures/alg-lexicase.tex
% TODO - In Algorithm 1, maybe this is too in the weeds but I might add "if not already evaluated" to 3.a, just because usually is in my implementations, so I don't even mention evaluation in the selection algorithm. I'd also change "less" to "worse" in 3.c to avoid specifying the +/- sign of goodness, which I think you avoid elsewhere.

\RestyleAlgo{boxruled}
\begin{algorithm}[h!]
    \begin{enumerate}
        \setlength\itemsep{1mm}

        \item Mark entire population as current \textbf{candidates} under consideration.
        
        \item Shuffle \textbf{test\_cases} into a random order.
    
        \item For each \textbf{case} in \textbf{test\_cases}:
            \begin{enumerate}
                \setlength\itemsep{1mm}
                \item Evaluate each candidate in \textbf{candidates} on \textbf{case}.
                
                \item Identify the \textbf{best\_score} on \textbf{case} of all candidates.
                
                \item Remove each entry from \textbf{candidates} with a score on \textbf{case} worse than \textbf{best\_score}.
                
            \end{enumerate}
                
            \item Select a random entry from \textbf{candidates}. 

    \end{enumerate}
 \caption{Lexicase selection for a single parent. Adapted from \citep{helmuth2015solving}.}
 \label{alg:lexicase}
\end{algorithm}

%% file: tex/diagnosing-lexicase-and-its-variants.tex
\section{Diagnosing the exploratory capacity of lexicase selection and its variants}
\label{sec:diagnosing-lexicase}

% -- Overview of what we did --
We conducted a series of experiments to analyze the exploratory limits of standard lexicase selection and four of its variants:
epsilon lexicase, down-sampled lexicase, cohort lexicase, and novelty-lexicase. 
For each experiment, unless stated otherwise, we evolved populations of 500 numerical vectors on our exploration diagnostic with a cardinality of 100 for 50,000 generations.
Across all experiments, we ran 50 replicates of each constituent treatment. 
We initialized populations to the lowest point in the fitness landscape, vectors of all 0.0s.

% -- Fitness assignment and selection --
When evaluating a candidate solution, we calculated a score associated with each position in its vector according to the exploration diagnostic (Figure~\ref{fig:genotype-phenotype}).
We used this collection of scores as test case qualities for lexicase selection and its variants.
For this work, we report quality directly; for comparison to other studies, note that test case \textit{error} is the amount that quality is below 100.
When a single fitness value was required (\textit{e.g.}, for tournament selection), we summed the individual test case qualities to determine the solution's aggregate fitness.

% -- Reproduction + mutation --
Selected candidate solutions reproduced asexually, and we applied point-mutations to offspring at a per-position rate of 0.7\%. 
The magnitude of each mutation was drawn from a normal distribution with a mean of 0.0 and a standard deviation of 1.0 ($\mathcal{N}(0, 1)$).
When mutations would raise a trait to a value $x$ where $x>100$, we rebounded that trait to $200-x$, ensuring that each trait value remained less than or equal to 100.
When mutations would lower a trait below 0.0, we reset that trait to 0.0.

% -- Analyses --
For each replicate of each experiment, we extracted the most performant individual in the population (\textit{i.e.}, the individual with the highest aggregate score) to compare across treatments.
For different diagnostic cardinalities (\textit{i.e.}, different numbers of test cases), the range of possible aggregate scores differs; as such, we normalized all aggregate scores by dividing by the cardinality, which results in a value between 0.0 and 100.0.

To identify the number of pathways being explored by a population, we measured the number of unique activation positions within each population.
Using this measurement, we calculated ``activation position coverage'' as the fraction of possible activation positions represented in a population.

% -- Software & data analysis -- 
For each experiment, we report both mean performance and mean activation position coverage over time (each with a bootstrapped 95\% confidence interval), and we compare measurements from the final generation across treatments.  
For each comparison, we performed a Kruskal-Wallis test to determine if there were significant differences; if so, we applied a Wilcoxon rank-sum test to distinguish between pairs of treatments, applying Bonferroni corrections for multiple comparisons where appropriate.

The software used to conduct experiments, statistical analyses, experimental data, and guides for replication are included in our supplemental material~\citep{supplemental_material}. 
See Section~\ref{sec:data-and-software-availability} for more details.

\subsection{Lexicase selection out-explores tournament selection}
\label{sec:diagnosing-lexicase:lexicase-vs-tournament}

\input{tex/figures/fig-tourny-vs-lexicase}

% Mini-methods
First, we used the exploration diagnostic to test well-established expectations that lexicase selection improves search space exploration relative to tournament selection.
Unlike lexicase selection, tournament selection does not reliably maintain multiple niches within a population~\citep{dolson_ecological_2018}; as such, we expected it to perform worse than lexicase selection on the exploration diagnostic.
For this experiment, we used tournaments of eight individuals. 

% Results
Consistent with our expectations, we found that lexicase selection outperforms tournament selection on the exploration diagnostic (Figure \ref{fig:tournament-vs-lexicase}; Wilcoxon rank-sum test: $p<10^{-4}$).
Early on, populations evolving under tournament selection converge to a single local optimum in the exploration diagnostic (\textit{i.e.}, a single activation position); without a mechanism to escape, populations become stuck and fail to continue exploring the search space.
Lexicase selection, however, rewards specialists for different activation positions, allowing the population to continuously explore different evolutionary pathways.
Indeed, we found that lexicase selection maintains substantially more ``activation-position'' specialists than tournament selection (Figure \ref{fig:tournament-vs-lexicase}; Wilcoxon rank-sum test: $p<10^{-4}$). 

\subsection{The exploratory capacity of lexicase selection degrades as we increase diagnostic cardinality}
\label{sec:diagnosing-lexicase:cardinality}

\input{tex/figures/fig-lexicase-cardinality}

Next, we evaluated standard lexicase selection on the exploration diagnostic at cardinalities 10, 20, 50, 100, 500, and 1,000.
Cardinality defines the number of potential pathways that must be explored by a population to guarantee finding the global optimum; increasing cardinality obscures the path to optimality.
Cardinality also corresponds to the number of test cases (\textit{i.e.}, niches) that individuals can specialize on. 
For a fixed population size, increasing the number of test cases decreases the long-term survival probability of any single specialist under lexicase selection~\citep{dolson_ecological_2018}, which could negatively affect lexicase's capacity to fully explore pathways in the search space.
For these reasons, we expected lexicase selection's performance on the exploration diagnostic to degrade as we increased cardinality.

Figure~\ref{fig:cardinality} shows lexicase selection's performance at each cardinality of the exploration diagnostic.
Across all cardinalities, lexicase selection improves performance over time.
Notably, treatments with cardinalities 10, 20, and 50 each perform near optimally after 50,000 generations, and populations evolved under cardinality 100 perform relatively well. 
Higher cardinalities (\textit{e.g.}, 200, 500, and 1000), however, perform substantially worse (Wilcoxon rank-sum tests: $p<10^{-4}$) and appear to need more time to converge on their maximal performance.
These data verify that increasing diagnostic cardinality also increases the exploration diagnostic's difficulty, as lexicase selection's performance degrades as cardinality increases.

We also found that populations evolved at lower diagnostic cardinalities maintained a larger coverage of unique activation positions than populations evolved at higher diagnostic cardinalities (Figure~\ref{fig:cardinality}).
Such diversity maintenance likely drove lexicase selection's ability to continuously explore pathways in the search space.

In these experiments, we used a population size of 500, resulting in 500 selection events per generation.
In each selection event, scores for vector positions (Figure~\ref{fig:genotype-phenotype}) are prioritized in a random order.
Across a population, we expect that positions that are consistently rewarded should maintain solutions that start at that position. 
The optimal solution requires the initial position to be the highest in the population, but this position may, by chance, never be evaluated first during lexicase selection.
The probability of this occurring varies with cardinality.
With a population size of 500 and a vector with 50 positions (\textit{i.e.}, a diagnostic cardinality of 50), there is a 0.004\% chance (1 in 25,000) of the initial position never being chosen first in a generation, making it unlikely to go unselected. 
Increasing the cardinality to 100, however, increases the chance for the first position to go unselected to 0.657\% (1 in 152)---a much more likely occurrence that may explain the reduced performance at cardinality 100 relative to cardinality 50.
By cardinality 200, the probability for the first position to go unselected within a given generation rises to 8.157\%, an even more likely occurrence.

One way to combat these dynamics is to increase population size, which would allow lexicase selection to support higher levels of exploration by reducing the chances of any given starting position from being skipped over by selection in any single generation.
However, increasing population size can be computationally expensive, as more individuals would need to be evaluated every generation.
Decreasing the depth of evolutionary search by reducing the number of generations evaluated is one way to balance the cost of increasing population size.
For a fixed computational budget, can increasing population size at the expense of evaluating fewer generations of evolution pay off under lexicase selection? 

\subsection{Increasing population size can improve lexicase selection's exploratory capacity}
\label{sec:diagnosing-lexicase:pop-size}

To test whether increasing population size can improve lexicase selection's exploratory capacity, we extended the runtime of our experiment and compared lexicase selection's performance on the exploration diagnostic (with a cardinality of 100) at two population sizes: 500 and 1,000. 
Because increasing population size increases per-generation computational effort, we ran both conditions for a fixed number of test case evaluations, evolving populations of 500 individuals for twice as many generations as populations of 1,000 individuals (1,000,000 and 500,000 generations, respectively).
As such, lineages from 500-individual populations take two reproductive steps in the search space for every one step reproductive step taken by a 1000-individual population.
This difference may allow the smaller populations to more rapidly exploit their initial position in the search space.
However, if larger populations are able to maintain more pathways in the search space, they may eventually outperform smaller populations. 

\input{tex/figures/fig-pop-size}

As expected, we found that increasing population size allows lexicase selection to maintain more starting positions for the entire duration of our experiment (Figure~\ref{fig:pop-size}).
Smaller populations initially outperform larger populations (given a fixed computational budget); however, despite running for fewer total generations, larger populations eventually outperform the smaller populations (Figure~\ref{fig:pop-size}; Wilcoxon rank-sum test: $p<10^{-4}$). 
These data suggest that, for a fixed number of test case evaluations, we can indirectly tune lexicase selection's level of search space exploitation and exploration by adjusting our allocation of computational resources between generations of evolution and population size.

% [Our result in context of previous program synthesis results: explaining/exploiting the advantages of down-sampled lexicase]
% [Most problems no significant difference in using larger populations versus evolving for more generations] % TODO - check to see if someone has done this
% [2 of 3 sampling rates of Scrabble score better at large population size, double letter and one of three of syllables better with more generations]
% [previous work done with down-sampled lexicase, so unclear if our result applies, but our result could suggest that scrabble score requires more exploration and double letter and syllables require more exploitation].

\subsection{Relaxing lexicase selection's elitism can improve exploration}
\label{sec:diagnosing-lexicase:epsilon-lexicase}

\input{tex/figures/fig-epsilon-lexicase}

As discussed in Section~\ref{sec:epsilon-lexicase-selection}, epsilon lexicase relaxes the elitism of lexicase selection.
To test whether this relaxation of elitism affects exploration, we compared standard lexicase selection and epsilon lexicase selection on the exploration diagnostic.
Specifically, we evolved 50 replicate populations at each of the following $\epsilon$ values: 0.0 (standard lexicase), 0.1, 0.3, 0.6, 1.2, 2.5, 5.0, and 10.0.

Epsilon lexicase with small values of $\epsilon$ (0.1 and 0.3) outperforms standard lexicase selection on the exploration diagnostic (Figure~\ref{fig:epsilon-lexicase}; Wilcoxon rank-sum tests: $p<10^{-4}$).
Extreme values of $\epsilon$ (5.0 and 10.0) significantly degrade performance relative to standard lexicase selection (Wilcoxon rank-sum tests: $p<10^{-4}$).
Interestingly, intermediate values of $\epsilon$ (0.6 and 1.2) perform best during the first approximately 20,000 generations, but are eventually outperformed by treatments with smaller values of $\epsilon$.
Unlike previous experiments, the relative levels of activation position coverage among conditions does not correspond with diagnostic performance.

% [relative amounts of starting position coverage does not predict relative performances among conditions].
% Standard lexicase and ...
% [chat about coverage results].

In general, epsilon lexicase is expected to have two main advantages over standard lexicase selection~\citep{la2016epsilon}: (1) it allows small amounts of noise in the evaluation data to be ignored, and (2) it prevents nearly identical scores from determining which candidate solutions win, potentially allowing for greater coexistence.
While the first mechanism cannot be at play here (since all scores are deterministic), the second advantage could provide additional support for solutions further along a given pathway. 
That is, solutions that begin optimizing at an earlier point in their vector, by definition, must have slightly lower values for later positions in their activated region.
In standard lexicase, when two solutions had overlapping activation regions, the one that start later would have an advantage at all overlapped sites.
In epsilon lexicase, however, the earlier start (\textit{i.e.}, the one with more long-term potential) now has a better chance to pass lexicase selection's selective filter. 
% with other solutions that began optimizing farther down in the vector and have a slightly higher value.
% @AML: The next two sentences were in the old version, but I wasn't totally sure what they were trying to say...
% [Eventually when a test case that favors the solutions that began optimizing from an earlier position could win and be selected as parents. This dynamic is neutralized if population diversity stagnates and too many similar solutions pass the same filters due to a higher $\epsilon$, leading to more randomness in selection.]

% Other potential $\epsilon$-lexicase configurations to pursue are using the an updated $\epsilon$ value, such as the median absolute deviation (MAD) \citep{la2016epsilon}. 
% This technique will update the $\epsilon$ value each generation, and is dependent on the fitness variance per individual test case.
% It is possible that this could lead to better performing solutions being found, given that the $\epsilon$ value will be continuously updated. 

\subsection{Down-sampling degrades lexicase selection's exploratory capacity}
\label{sec:diagnosing-lexicase:down-sampled-lexicase}

\input{tex/figures/fig-down-sampled-lexicase}

Next, we investigated whether down-sampling affects lexicase selection's exploratory capacity by comparing the performance of lexicase selection at a range of sampling rates: 100\% (standard lexicase), 50\%, 20\%, 10\%, 5\%, 2\%, and 1\%. 
For example, a 10\% sampling rate means that each generation we randomly selected 10 of the 100 possible test cases (for a diagnostic cardinality of 100) to be used for parent selection.
Down-sampling reduces the per-generation computational effort required for parent selection by conducting fewer test case evaluations (Section~\ref{sec:down-sampled-lexicase-selection}).
For a fair comparison across different sampling rates, we limited the computational budget to a maximum of $2.5\times10^9$ test case \textit{evaluations} by varying the number of generations of evolution for each subsampling rate (100\%: 50,000 generations,
50\%: 100,000 generations, 
20\%: 250,000 generations, 
10\%: 500,000 generations, 
5\%: 1,000,000 generations, 
2\%: 2,500,000 generations, 
and 
1\%: 5,000,000 generations). 

Any amount of down-sampling significantly degraded lexicase selection's performance on the exploration diagnostic for the allotted computational budget (Figure~\ref{fig:down-sampled-lexicase}; Wilcoxon-rank sum tests: $p<10^{-4}$).
Down-sampled lexicase selection's drop in performance is likely attributed to frequent mismatches between candidate solutions and the test cases that they are specialized on.
As the proportion of test cases used in each generation decreases, so too does the probability of a solution encountering the same set of test cases for multiple generations in a row. 
As such, a solution has a reduced chance of encountering the test cases for which it is most optimized~\citep{ferguson2020characterizing}.
These dynamics will repeatedly remove solutions with small active regions, thereby reducing population diversity.
Indeed, we found that down-sampling substantially reduces the number of activation position specialists represented in the population (Figure~\ref{fig:down-sampled-lexicase}; Wilcoxon rank-sum tests: $p<10^{-4}$).
In fact, any down-sampling used appears to have a strong negative effect, substantially reducing performance in all cases.

% @AML: this paragraph might be a little rambling... 
% A major reason for deciding to use down-sampled lexicase instead of standard lexicase selection is to reduce the number of test case evaluations needed for parent selection, which can allow for greater population sizes or more generations of evolution for a fixed evaluation budget [CITE].
% Given that we initialized populations to the lowest position in the search space (vectors of all zeroes), the beginning of each run is likely dominated by rapid exploitation, as populations rapidly exploit the evolutionary pathways sampled by early mutations.
% Thus, early on, rapid generational turnover (\textit{i.e.}, deeper evolutionary search) is ideal, as initial pathways in the search space can be rapidly explored.
% Indeed, lexicase selection with a moderate amount of down-sampling initially outperforms standard lexicase (Figure~\ref{alg:down-sampled-lexicase}); however, standard lexicase selection eventually overtakes all forms of down-sampled lexicase, as continual exploration of different pathways in the search space becomes increasingly important for continued optimization.

% @AML: !! What does this say about our hypothesis of increased generational turnover?? 
We repeated this experiment, except we increased population size instead of increasing generations of evolution for down-sampled lexicase; that is, we ran each condition for an equivalent number of generations but differing population sizes to maintain a fixed number of evaluations.
We report these data in our supplemental material~\citep{supplemental_material}.
Overall, the patterns were similar to that of increasing generations of evolution.
Initially, down-sampled lexicase outperforms standard lexicase on the exploration diagnostic; however, standard lexicase eventually outperforms down-sampled lexicase across all subsampling rates~\citep{supplemental_material}.

\subsection{Cohort partitioning degrades lexicase selection's exploratory capacity}
\label{sec:diagnosing-lexicase:cohort-lexicase}

\input{tex/figures/fig-cohort-lexicase}

Next, we evaluated whether partitioning the population and test cases into cohorts affects the exploration capacity of lexicase selection.
We compared the performance of standard lexicase to that of cohort lexicase at a range of cohort sizes (given as the proportion of the population and the set of test cases used in each cohort): 
100\% (standard lexicase), 50\%, 20\%, 10\%, 5\%, 2\%, and 1\%.
For example, a cohort size of 10\% means that the population (of 500 individuals) is divided into 10 cohorts of 50 individuals each, and the test cases (100 total) are also divided into those same 10 cohorts, with 10 test cases in each. 
Like down-sampled lexicase, cohort lexicase reduces the per-generation computational effort required for parent selection by evaluating each cohort of candidate solutions on only one of the test case cohorts (Section~\ref{sec:cohort-lexicase-selection}).
Likewise, for fair comparison across different cohort sizes, we limited the computational budget to a maximum of $2.5\times10^9$ test case evaluations by varying the number of generations of evolution for each cohort size (100\%: 50,000 generations,
50\%: 100,000 generations, 
20\%: 250,000 generations, 
10\%: 500,000 generations, 
5\%: 1,000,000 generations, 
2\%: 2,500,000 generations, 
and 
1\%: 5,000,000 generations). 

As with down-sampled lexicase, any level of cohort partitioning degrades lexicase's performance on the exploration diagnostic for the allotted computational budget (Figure~\ref{fig:cohort-lexicase}; Wilcoxon rank-sum tests: $p<10^{-4}$).
However, cohort lexicase does not appear to degrade lexicase selection's performance to the same degree as down-sampled lexicase for a given subsampling rate (Figure~\ref{fig:down-sampled-lexicase}).
Moreover, standard lexicase took longer (more total evaluations) to outperform cohort lexicase than to outperform down-sampled lexicase. 
These data suggest that cohort partitioning (with intermediate levels of partitioning) may be a better method of random subsampling in the context of lexicase selection. 

% a more appropriate parent selection algorithm to use instead of down-sampled lexicase given a limited computational budget and a problem that requires a high degree of search space exploration to solve.]

We repeated this experiment, except we increased population size instead of increasing generations of evolution for cohort lexicase; that is, we ran each condition for an equivalent number of generations but differing population sizes to maintain a fixed number of evaluations.
We report these data in our supplemental material~\citep{supplemental_material}.
The overall patterns were qualitatively different and warrant further exploration in future work. 
We found no compelling evidence that cohort lexicase outperformed standard lexicase in the given computational budget; however, we did find that populations evolving under cohort lexicase (with larger population sizes) maintained more activation position coverage than standard lexicase selection~\citep{supplemental_material}.
Further, some of the cohort sizes were on an upward trajectory when the runs finished and may eventually outperform standard lexicase given a larger computational budget. 

% Neither cohort nor down-sampled lexicase performed well at small cohort and sample sizes; both lexicase variants failed to maintain 
% As with down-sampled lexicase with small sample sizes, cohort lexicase  small cohort sizes, cohort lexicase performs 

\subsection{Cohort lexicase out-explores down-sampled lexicase}
\label{sec:diagnosing-lexicase:cohort-vs-down-sampled}

\input{tex/figures/fig-down-sampled-vs-cohort}

Next, we independently verified that cohort lexicase out-explores down-sampled lexicase on the exploration diagnostic.
To do so, we compared the performance of cohort lexicase and down-sampled lexicase with their most performant parameterizations: a 50\% cohort size and a 50\% sampling rate, respectively.
We again limited the computational budget to a maximum of $2.5\times10^9$ test case evaluations (100,000 generations of evolution for both conditions), and we ran 50 new replicates of each condition for comparison.

% IDEA for another control: compare to island model systems
%  - two runs of half the population size vs one run of full population size
%  - cohort structure independent of sampling

As expected given Figures~\ref{fig:down-sampled-lexicase} and~\ref{fig:cohort-lexicase}, cohort lexicase outperformed down-sampled lexicase by a substantial margin for the given computational budget (Figure~\ref{fig:down-sampled-vs-cohort}; Wilcoxon rank-sum test: $p<10^{-4}$). 
Interestingly, down-sampled lexicase appears to briefly outperform cohort lexicase in the first few thousand generations but is quickly overtaken by cohort lexicase. 
Both cohort and down-sampled lexicase offer equivalent per-generation evaluation savings, but cohort lexicase uses every test case for parent selection in every generation.
This could play a role in problem-solving success, as a test case that rewards exploration at any given activation position in the exploration diagnostic is used every generation.
Indeed, populations evolving under cohort lexicase selection maintained a higher diversity of activation positions than populations evolving under down-sampled lexicase selection (Figure~\ref{fig:down-sampled-vs-cohort}; Wilcoxon rank-sum test: $p<10^{-4}$).

Previous work predicted the potential for such differences between cohort and down-sampled lexicase. 
\cite{ferguson2020characterizing} found that cohort lexicase better maintained phylogenetic diversity than down-sampled lexicase, as phylogenies coalesced less frequently under cohort lexicase selection (maintaining deeper, more divergent branches).
Despite this difference in diversity maintenance, \cite{ferguson2020characterizing} did not find significant differences in problem-solving success across a set of program synthesis benchmark problems, which suggests that the test cases used in these benchmark problems were more robust to random subsampling than the test cases for the exploration diagnostic.
Indeed, each individual test case for the exploration diagnostic uniquely represents a single activation position; that is, test cases are minimally redundant with one another.
In many program synthesis benchmark problems, however, individual test cases are often intentionally redundant to others, differing only in the particular values of their inputs and outputs and not necessarily different in the functional specialization they reward. 
Such redundancies prevent candidate solutions from memorizing particular input-output pairings, forcing candidate solutions to generalize in order to achieve high fitness across redundant test cases.
This detail could explain why the exploration diagnostic reveals substantial performance differences between cohort and down-sampled lexicase where more standard benchmark problems failed to do so.

\subsection{Novelty test cases degrade lexicase selection's exploratory capacity}
\label{sec:diagnosing-lexicase:novelty-lexicase}

\input{tex/figures/fig-novelty-lexicase}

Finally, we evaluated how incorporating novelty test cases into lexicase selection impacts exploration.
We compared the performance of standard lexicase to that of novelty-lexicase for a range of $k$-nearest neighbors: 0 (standard lexicase), 1, 2, 4, 8, 15, 30, and 60.

Contrary to our expectations, we found that the addition of novelty test cases degrades performance on the exploration diagnostic in all cases (Figure~\ref{fig:novelty-lexicase}; Wilcoxon rank-sum test: $p<10^{-4}$).
Though, novelty-lexicase generally maintains similar levels of activation position diversity in the population relative to standard lexicase, and by the end of the experiment, some parameterizations of novelty lexicase maintain more activation positions, though none of the differences appear to be substantial (Figure~\ref{fig:novelty-lexicase}). 

Novelty search favors solutions that have never been seen before, regardless of their impact on fitness. 
Based on previous studies, we expected novelty-lexicase to outperform standard lexicase on the exploration diagnostic~\citep{jundt_comparing_2019}.
However, novelty-lexicase appears to hinder lexicase's ability to fully exploit pathways in the diagnostic's search space.

While past work has demonstrated that novelty search can be effective at producing solutions for complicated problems, the exploration diagnostic does not have any of the hidden intricacies that novelty search excels at disentangling.
Indeed, novelty search appears to thrive under conditions where there are more non-linearities between genotype and phenotype.
The underlying representation used here is purposely simple numerical vectors as opposed to an artificial neural network~\citep{lehman_exploiting_2008} or PushGP~\citep{jundt_comparing_2019} where internal architectures can change and qualitatively different outputs are possible.
For example, in this case, all sites in a genome are optimal at one end of their range of values, whereas most complex problems are assumed to have pockets of solutions throughout the genotype-phenotype map.
Additionally, our results also used a single, limited form of novelty lexicase.  
We did not use a seed bank (the importance of which has previously been stressed), and we used k-nearest neighbors euclidean distances to measure novelty instead of a direct measure of behavioral uniqueness.
These differences in problems may shine a light as to why novelty-lexicase did not outperform standard lexicase selection on the exploration diagnostic.

Our results from varying diagnostic cardinality (Section~\ref{sec:diagnosing-lexicase:cardinality}) may also offer insights into the unexpectedly poor performance of novelty-lexicase selection.
Novelty-lexicase selection increases the number of test cases used for parent selection (in this work, doubling the number of test cases from 100 to 200).
Increasing the number of test cases (without simultaneously increasing the population size) is not without cost, degrading specialist maintenance and performance on the exploration diagnostic (Figure~\ref{fig:cardinality}).
This dynamic is likely to be at play in our novelty-lexicase experiment, as population size was constant for both standard lexicase and novelty-lexicase selection.
% In fact, standard lexicase's performance relative to novelty-lexicase is all the more impressive given that it required half as many test case evaluations per generation, thus using less computational resources overall.

%% file: tex/figures/fig-tourny-vs-lexicase.tex
\begin{figure}[ht!]
\centering
\includegraphics[width=\textwidth]{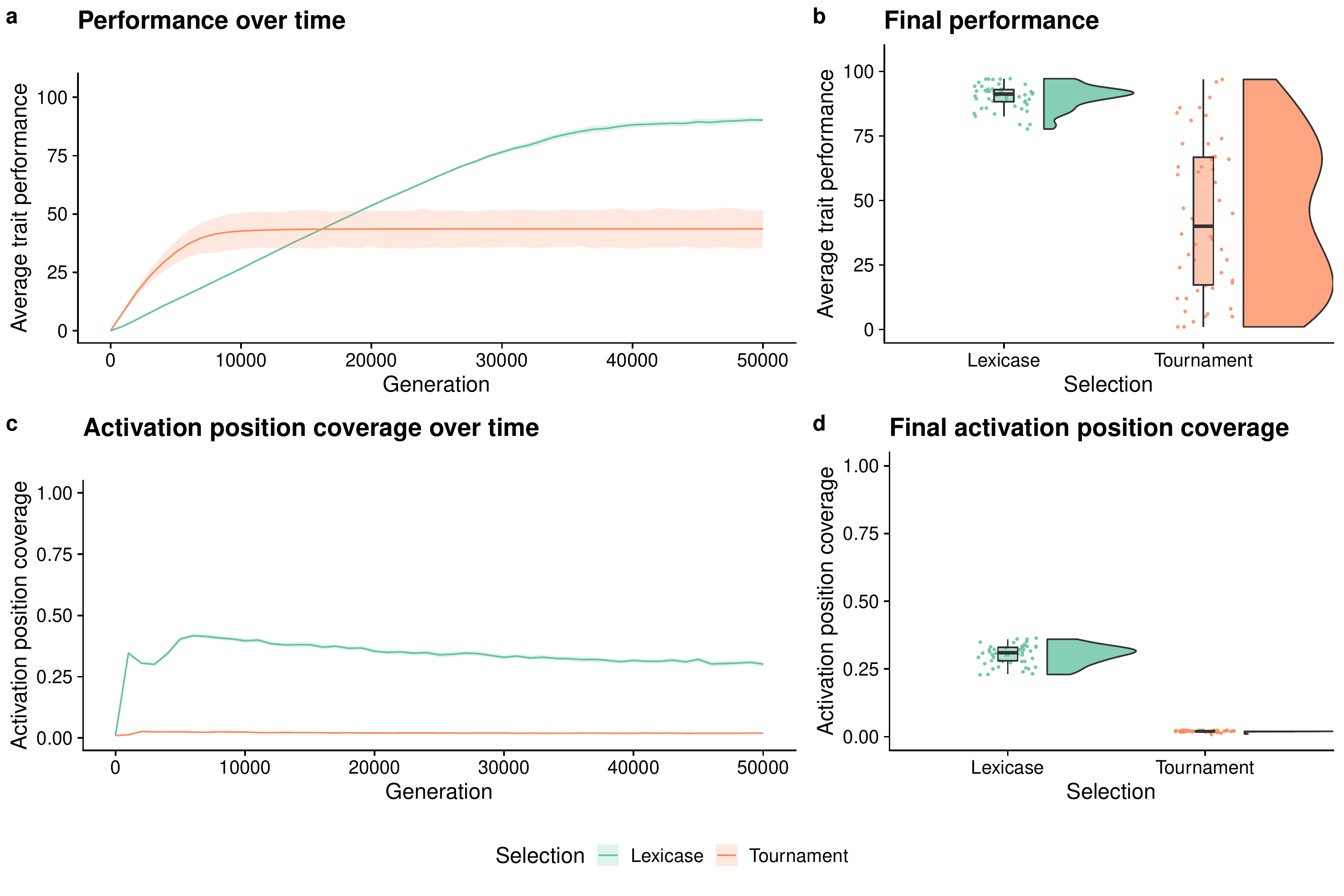}
\caption{
\textbf{Lexicase selection versus tournament selection on the exploration diagnostic.} 
Panels (a) and (b) show performance over time and at the end of 50,000 generations, respectively.
Likewise, panels (c) and (d) show activation position coverage over time and at the end of 50,000 generations, respectively.
For panels (a) and (c), each line gives the mean value across 50 replicates, and the shading around each mean gives a 95\% confidence interval.
% Panels (b) and (d) are annotated with statistically significant relationships (Wilcoxon rank-sum test).
}
\label{fig:tournament-vs-lexicase}  
\end{figure}

%% file: tex/figures/fig-lexicase-cardinality.tex
\begin{figure}[ht!]
\centering
\includegraphics[width=\textwidth]{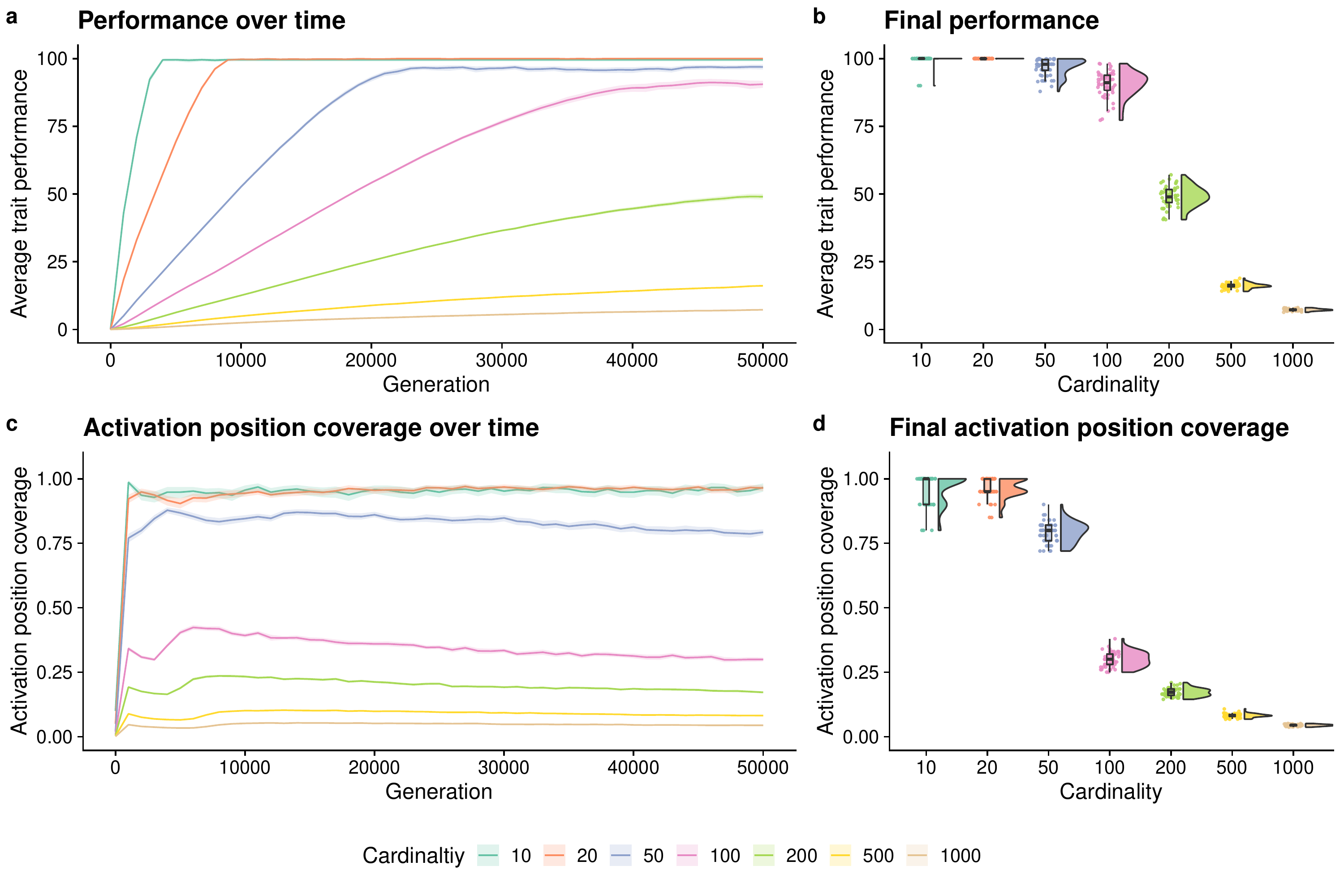}
\caption{
\textbf{Lexicase selection at a range of exploration diagnostic cardinalities.} 
Panels (a) and (b) show performance over time and at the end of 50,000 generations, respectively.
Likewise, panels (c) and (d) show activation position coverage over time and at the end of 50,000 generations, respectively.
For panels (a) and (c), each line gives the mean value across 50 replicates, and the shading around each mean gives a 95\% confidence interval.
}
\label{fig:cardinality}
\end{figure}

%% file: tex/figures/fig-pop-size.tex
\begin{figure}[ht!]
\centering
\includegraphics[width=\textwidth]{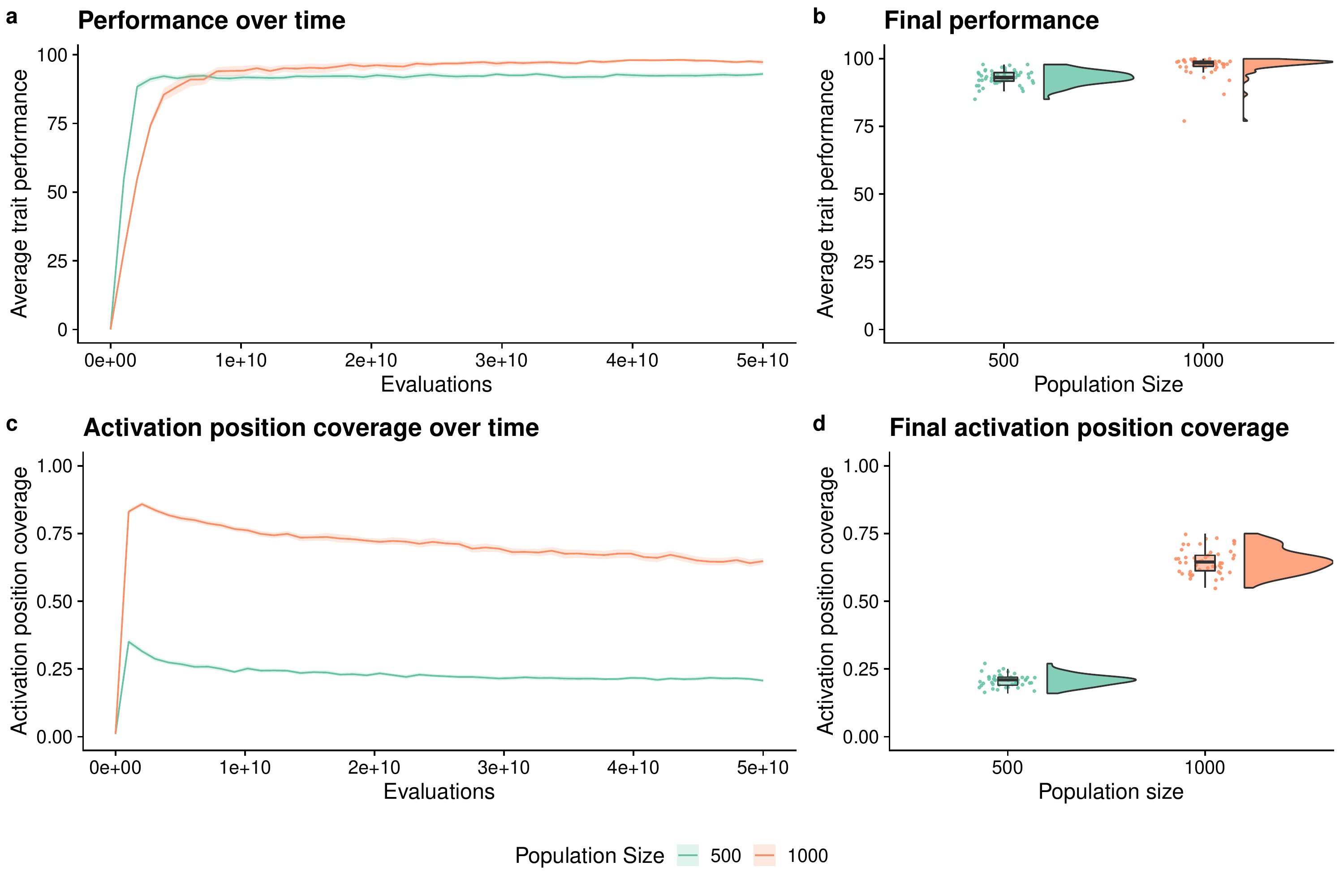}
\caption{\textbf{Lexicase selection's performance on the exploration diagnostic at different population sizes.} 
Panels (a) and (b) show performance over time and at the end of the experiment, respectively.
Likewise, panels (c) and (d) show activation position coverage over time and at the end of the experiment, respectively.
For panels (a) and (c), each line gives the mean value across 50 replicates, and the shading around each mean gives a 95\% confidence interval.
}
\label{fig:pop-size}  
\end{figure}

%% file: tex/figures/fig-epsilon-lexicase.tex
\begin{figure}[ht!]
\centering
\includegraphics[width=\textwidth]{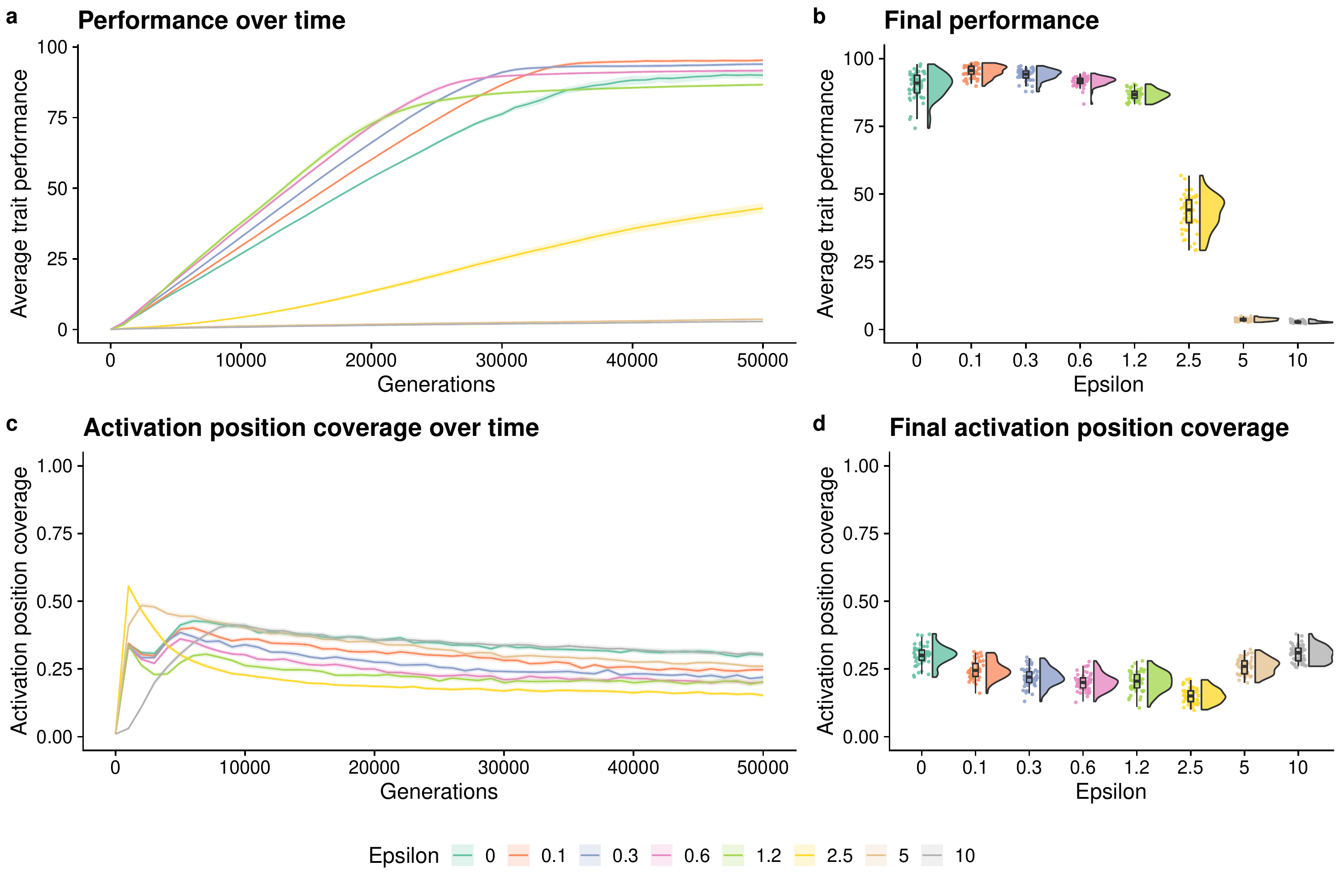}
\caption{
\textbf{Epsilon lexicase selection's performance on the exploration diagnostic at a range of $\epsilon$ values.} 
Panels (a) and (b) show performance over time and after 50,000 generations of evolution, respectively.
Likewise, panels (c) and (d) show activation position coverage over time and after 50,000 generations of evolution, respectively.
For panels (a) and (c), each line gives the mean value across 50 replicates, and the shading around each mean gives a 95\% confidence interval.
}
\label{fig:epsilon-lexicase}
\end{figure}

%% file: tex/figures/fig-down-sampled-lexicase.tex
\begin{figure}[ht!]
\centering
\includegraphics[width=\textwidth]{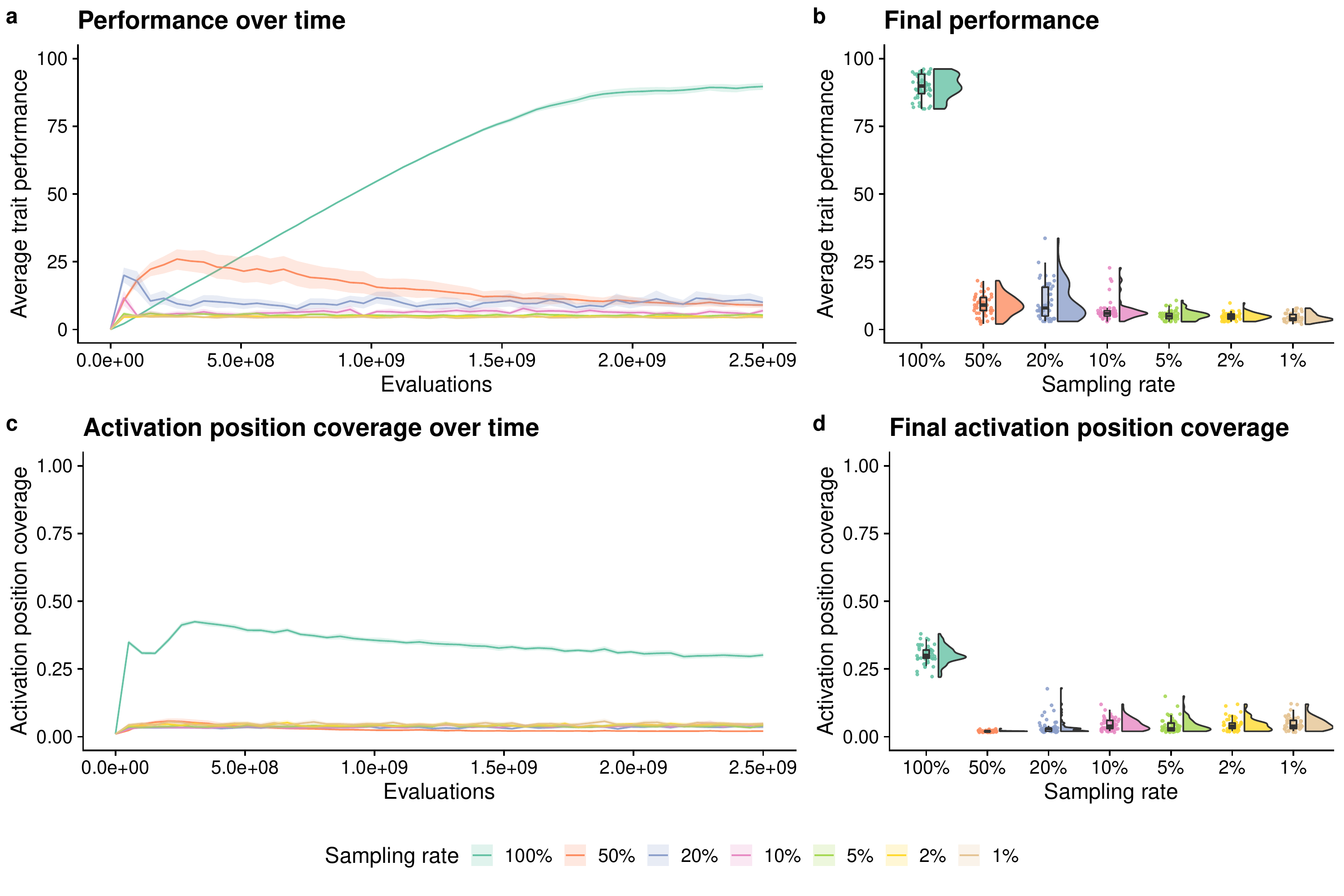}
\caption{
\textbf{Down-sampled lexicase selection's performance on the exploration diagnostic at a range of subsampling rates.}
Panels (a) and (b) show performance over time and at the end of the experiment, respectively.
Likewise, panels (c) and (d) show activation position coverage over time and at the end of the experiment, respectively.
For panels (a) and (c), each line gives the mean value across 50 replicates, and the shading around each mean gives a 95\% confidence interval.
}
\label{fig:down-sampled-lexicase}
\end{figure}

%% file: tex/figures/fig-cohort-lexicase.tex
\begin{figure}[ht!]
\centering
\includegraphics[width=\textwidth]{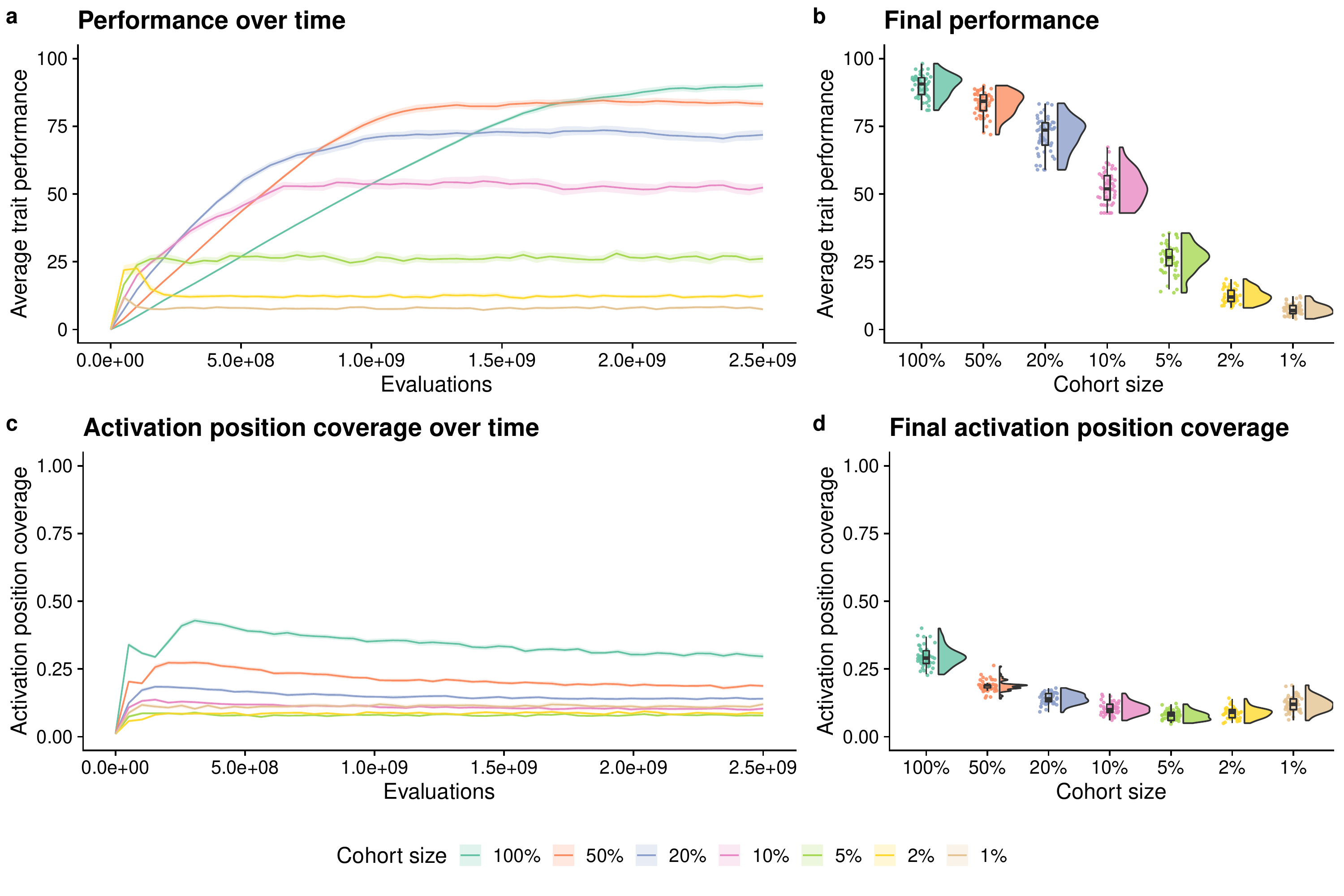}
\caption{
\textbf{Cohort lexicase selection's performance on the exploration diagnostic at a range of partitioning rates.}
Panels (a) and (b) show performance over time and at the end of the experiment, respectively.
Likewise, panels (c) and (d) show activation position coverage over time and at the end of the experiment, respectively.
For panels (a) and (c), each line gives the mean value across 50 replicates, and the shading around each mean gives a 95\% confidence interval.
} 
\label{fig:cohort-lexicase}
\end{figure}

%% file: tex/figures/fig-down-sampled-vs-cohort.tex
\begin{figure}[ht!]
\centering
\includegraphics[width=\textwidth]{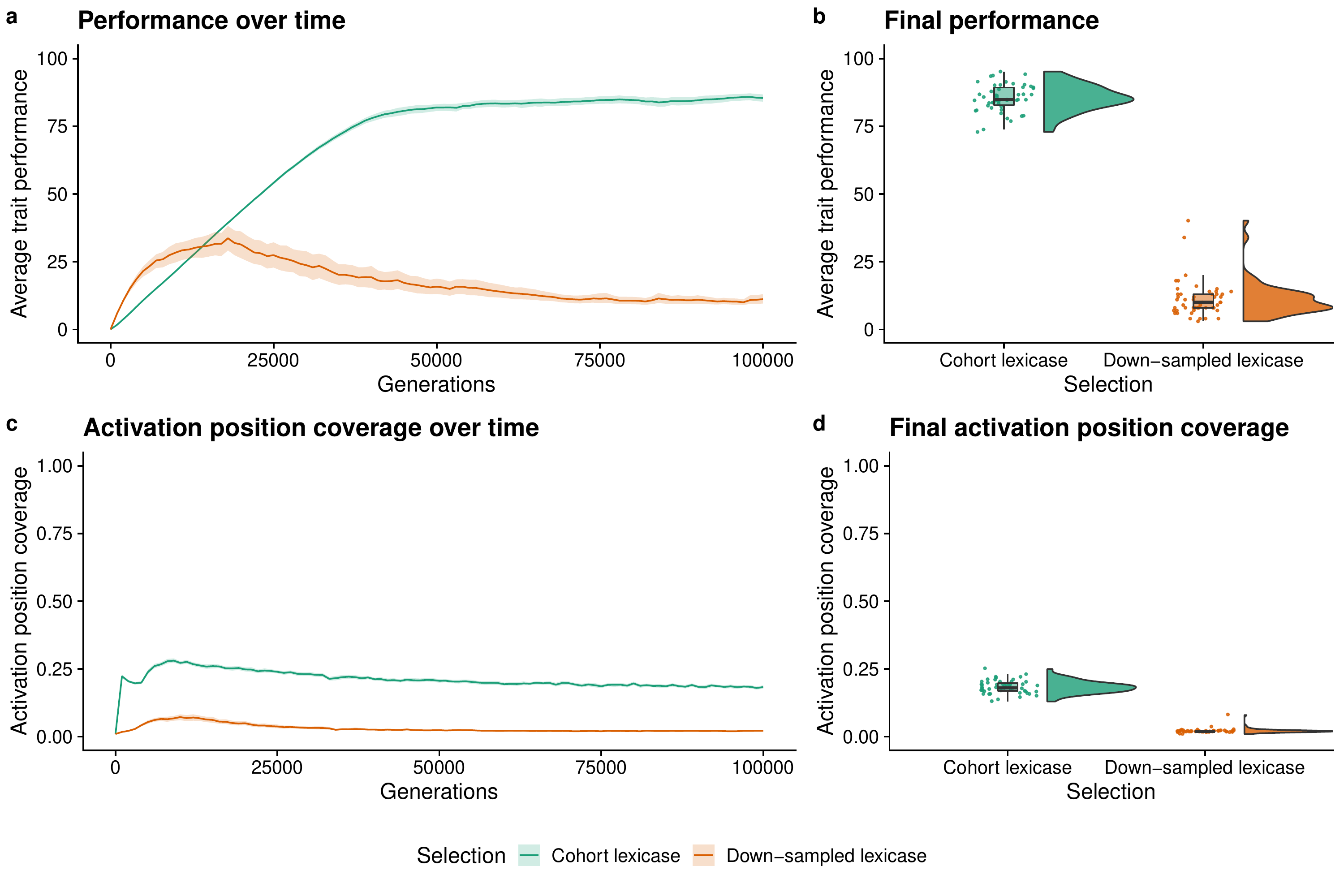}
\caption{
\textbf{Down-sampled versus cohort lexicase on the exploration diagnostic.} 
Panels (a) and (b) show performance over time and at the end of the experiment, respectively.
Likewise, panels (c) and (d) show activation position coverage over time and at the end of the experiment, respectively.
For panels (a) and (c), each line gives the mean value across 50 replicates, and the shading around each mean gives a 95\% confidence interval.
}
\label{fig:down-sampled-vs-cohort}
\end{figure}

%% file: tex/figures/fig-novelty-lexicase.tex
\begin{figure}[ht!]
\centering
\includegraphics[width=\textwidth]{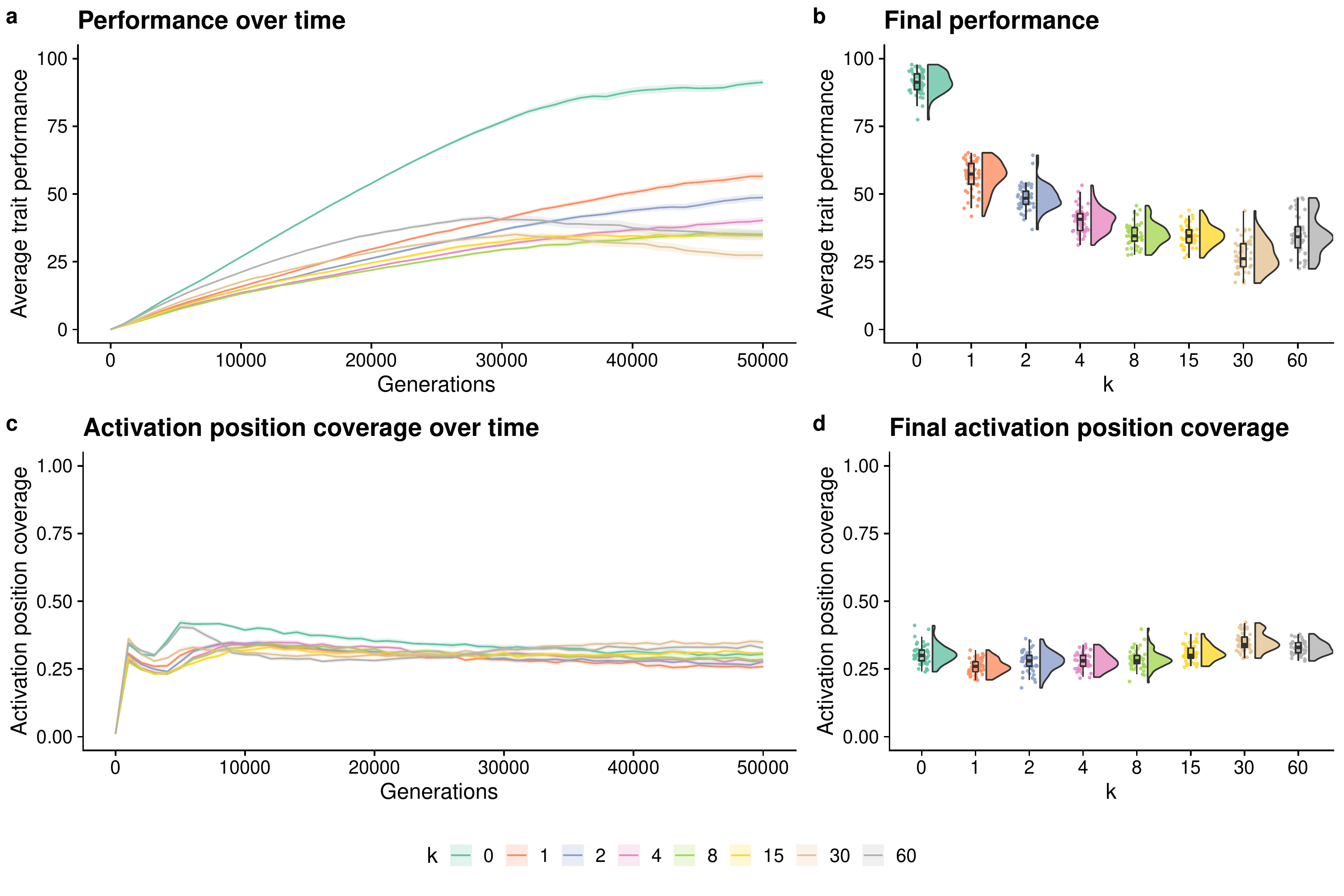}
\caption{
\textbf{Novelty-lexicase selection's performance on the exploration diagnostic at a range of nearest-neighbor parameterizations.} 
Panels (a) and (b) show performance over time and after 50,000 generations of evolution, respectively.
Likewise, panels (c) and (d) show activation position coverage over time and after 50,000 generations of evolution, respectively.
For panels (a) and (c), each line gives the mean value across 50 replicates, and the shading around each mean gives a 95\% confidence interval.
}
\label{fig:novelty-lexicase}
\end{figure}

%% file: tex/conclusion.tex
\section{Conclusion}
\label{sec:conclusion}

% TODO - Does the fact that dependency structure among cases (= loci) is linear matter? It seems to me that it might. What if it was a DAG?

% -- Summary of what we did --
% In this work, we introduced an  
% We used this exploration diagnostic to investigate the exploratory limits of lexicase selection along with several of its variants: epsilon lexicase, down-sampled lexicase, cohort lexicase, and novelty-lexicase.

In this work, we introduced a new diagnostic to investigate the exploratory limits of lexicase selection along with several of its variants: epsilon lexicase, down-sampled lexicase, cohort lexicase, and novelty-lexicase.
First, we verified well-established expectations that lexicase selection better facilitates search space exploration than tournament selection. 
Across all exploration diagnostic difficulty levels (\textit{i.e.}, cardinalities), lexicase selection drove improvements in performance (Figure~\ref{fig:cardinality}), while tournament selection repeatedly failed to escape early local optima (Figure~\ref{fig:tournament-vs-lexicase}).
As we increased the cardinality of the diagnostic, lexicase selection's specialist maintenance and overall performance waned.
Conditions with larger diagnostic cardinalities used more test cases to evaluate individuals, and as such had more possible specialists (\textit{i.e.}, niches).
Given a fixed population size, lexicase maintained a smaller fraction of possible specialists as the number of possible niches increased, which, in turn, decreased overall performance (Figure~\ref{fig:cardinality}).

Interestingly, we found that allocating a computational budget (\textit{i.e.}, candidate solution evaluations) toward increasing generations versus increasing population size is not necessarily a straightforward choice when using lexicase selection. 
In our case, a larger population size enabled better specialist maintenance and ultimately higher performance on the exploration diagnostic with standard lexicase (Figure~\ref{fig:pop-size}).
This finding is interesting in light of \cite{helmuth_problem-solving_2021}'s work investigating the problem-solving benefits of down-sampled lexicase; on a suite of program synthesis problems, Helmuth and Spector found that some problems benefited from an increased population size (at the cost of running for fewer generations), some problems benefited from an increase in generations, and most problems were unaffected by their choice of increasing population size versus generations evaluated. 

% -- Lexicase selection is sensitive to training set size --
Overall, these results suggest that lexicase selection can be sensitive to expanding the set of test cases used for evaluation, especially if each test case uniquely represents a distinct, desirable trait. 
Moreover, our results suggest the importance of more deeply examining the benchmark problems that we use and the characteristics of the search spaces that they represent. 
Given a fixed computational budget, why do some problems benefit from running deeper evolutionary searches while others benefit from increased population sizes under lexicase selection?
For many problems, different categories of test cases have uneven representation in the test set.
We hypothesize that the distribution of test cases among categories plays a role in lexicase selection's success and the optimal balance between population size and depth of search (generations of evolution). 
For example, if the number of test cases is similar to population size, lexicase selection may fail to maintain specialists on categories that are underrepresented in the test cases and instead favor overrepresented categories.
% In many program synthesis benchmark problems, individual test cases are often highly redundant to others, differing only in the particular values of their inputs and outputs and not necessarily in the functional specialization they reward. 
In future work, we will develop novel diagnostic tools for investigating the sensitivity of selection schemes to test case set composition.

% -- Variations on standard lexicase affect exploration --
We found that each of the lexicase variants that we evaluated---epsilon lexicase, down-sampled lexicase, cohort lexicase, and novelty-lexicase---affected lexicase selection's exploratory capacity.
For small values of $\epsilon$, epsilon lexicase outperformed standard lexicase selection on the exploration diagnostic, while large values of $\epsilon$ substantially degraded performance.
Surprisingly, we found that novelty-lexicase degrades performance on the exploration diagnostic relative to standard lexicase selection.

% We offer two, non-mutually exclusive potential mechanisms for novelty-lexicase poor diagnostic performance to be explored by future work: the exploration diagnostic's lack of a complex genotype-phenotype mapping and novelty-lexicase dilution of test cases]

% -- down-sampled & cohort lexicase --
Our experiments are also the first to demonstrate consequential differences between down-sampled and cohort lexicase selection, as previous work generally failed to distinguish the problem-solving performance of these two lexicase variants~\citep{ferguson2020characterizing}.
Cohort lexicase substantially outperformed down-sampled lexicase (Figure~\ref{fig:down-sampled-vs-cohort}).
Both down-sampled and cohort lexicase offer equivalent per-generation evaluation savings, so our results suggest that cohort partitioning may often be a better subsampling method than down-sampling for lexicase selection.
Future work should examine whether this difference between cohort partitioning and down-sampling holds across different selection schemes.

Given equivalent computational budgets, we found that standard lexicase selection eventually outperforms both cohort and down-sampled lexicase on the exploration diagnostic (Figures~\ref{fig:down-sampled-lexicase} and~\ref{fig:cohort-lexicase}).
This result diverges from recent benchmarking studies where subsampling substantially improved performance on a range of program synthesis problems~\cite{ferguson2020characterizing,helmuth_explaining_2020,helmuth_problem-solving_2021}.
Future work will develop diagnostic problems to help identify when subsampling (\textit{e.g.}, via either cohort partitioning or down-sampling) is likely to improve versus impede lexicase selection's performance.

% -- overall, future work --
In each of our experiments, we focused our analyses on performance and activation position diversity maintenance.
Future work should more deeply examine the evolutionary histories of evolving populations using phylodiversity metrics~\citep{dolson_interpreting_2020}.
Along with this, other parameter values and configurations of each of the variants evaluated in this work could be tested in order to develop a more complete understanding of how parameterization affects exploration.

We intend for this work to demonstrate how diagnostics (\textit{e.g.}, the exploration diagnostic introduced here) can be valuable tools for evaluating the pros and cons of different selection schemes.
We plan to implement a larger suite of selection scheme diagnostics, each targeted toward evaluating a particular aspect of problem-solving.
Such diagnostics will complement conventional benchmarking experiments in our community's effort to understand how different selection schemes steer evolutionary search.

% Does the fact that dependency structure among cases (= loci) is linear matter? It seems to me that it might. What if it was a DAG?
% This exploration diagnostic can also be used to compare the exploratory capabilities of different selection schemes.

%% file: tex/software-data-availability.tex
\section{Data and Software Availability}
\label{sec:data-and-software-availability}

Our supplemental material~\citep{supplemental_material} is hosted on \href{https://github.com/jgh9094/GPTP-2021-Exploration-Of-Exploration}{GitHub} and contains the software, data analyses, and documentation associated with this work.
Our experiments are implemented using the Empirical library~\citep{charles_ofria_2020_empirical}, and we used a combination of Python and R version 4 \citep{r_lang} for data processing and analysis.
We used the following R packages for data wrangling, statistical analysis, graphing, and visualization: 
ggplot2 \citep{R-ggplot2}, 
tidyverse \citep{R-tidyverse}, 
knitr \citep{R-knitr}, 
cowplot \citep{R-cowplot}, 
viridis \citep{R-viridis}, 
RColorBrewer \citep{R-RColorBrewer}, 
rstatix \citep{R-rstatix}, 
ggsignif \citep{R-ggsignif}, 
Hmisc \citep{R-Hmisc}, 
and 
kableExtra \citep{R-kableExtra}.
We used R markdown \citep{R-markdown} and bookdown \citep{R-bookdown} to generate web-enabled supplemental material.
Our experimental data is available on the Open Science Framework at \url{https://osf.io/xpjft/} \citep{supplemental_data}.

%% file: tex/acknowledgements.tex
\begin{acknowledgement}
 
% Acknowledgements:
%  - Spector for encouraging remarks and feedback on the manuscript
%  - GPTP participants for lively discussion and feedback?
%  - Devolab
%  - Hpcc
%  - Funding sources

We thank members of the Michigan State University (MSU) Digital Evolution Laboratory for helpful comments and suggestions on this work.
We thank the participants of the 2021 Genetic Programming in Theory and Practice workshop for lively discussion of our work.  
We especially thank Lee Spector for encouraging remarks and insightful feedback on our manuscript.
MSU provided computational resources through the Institute for Cyber-Enabled Research.
This work was supported in part by the National Science Foundation (NSF) through the BEACON Center (DBI-0939454) and a Graduate Research Fellowship to AL (DGE-1424871) and by the GEM Fellowship Program and Oak Ridge National Laboratory (ORNL).
Any opinions, findings, and conclusions or recommendations expressed in this material are those of the author(s) and do not necessarily reflect the views of MSU, the NSF, GEM, or ORNL. 

\end{acknowledgement}

%% file: main.bbl
\begin{thebibliography}{}

\bibitem[Aenugu and Spector, 2019]{aenugu_lexicase_2019}
Aenugu, S. and Spector, L. (2019).
\newblock Lexicase selection in learning classifier systems.
\newblock In {\em Proceedings of the {Genetic} and {Evolutionary} {Computation}
  {Conference} on - {GECCO} '19}, pages 356--364, Prague, Czech Republic. ACM
  Press.

\bibitem[Ahlmann-Eltze and Patil, 2021]{R-ggsignif}
Ahlmann-Eltze, C. and Patil, I. (2021).
\newblock {\em ggsignif: Significance Brackets for ggplot2}.
\newblock R package version 0.6.2.

\bibitem[Allaire et~al., 2020]{R-markdown}
Allaire, J., Xie, Y., McPherson, J., Luraschi, J., Ushey, K., Atkins, A.,
  Wickham, H., Cheng, J., Chang, W., and Iannone, R. (2020).
\newblock {\em rmarkdown: Dynamic Documents for R}.
\newblock R package version 2.6.

\bibitem[Dolson et~al., 2020]{dolson_interpreting_2020}
Dolson, E., Lalejini, A., Jorgensen, S., and Ofria, C. (2020).
\newblock Interpreting the {Tape} of {Life}: {Ancestry}-{Based} {Analyses}
  {Provide} {Insights} and {Intuition} about {Evolutionary} {Dynamics}.
\newblock {\em Artificial Life}, 26(1):58--79.

\bibitem[Dolson et~al., 2018]{dolson_ecological_2018}
Dolson, E.~L., Banzhaf, W., and Ofria, C. (2018).
\newblock Ecological theory provides insights about evolutionary computation.
\newblock preprint, PeerJ Preprints.

\bibitem[Eiben and Schippers, 1998]{eiben1998evolutionary}
Eiben, A.~E. and Schippers, C.~A. (1998).
\newblock On evolutionary exploration and exploitation.
\newblock {\em Fundamenta Informaticae}, 35(1-4):35--50.

\bibitem[Ferguson et~al., 2020]{ferguson2020characterizing}
Ferguson, A.~J., Hernandez, J.~G., Junghans, D., Lalejini, A., Dolson, E., and
  Ofria, C. (2020).
\newblock Characterizing the effects of random subsampling on lexicase
  selection.
\newblock In {\em Genetic Programming Theory and Practice XVII}, pages 1--23.
  Springer.

\bibitem[Garnier, 2018]{R-viridis}
Garnier, S. (2018).
\newblock {\em viridis: Default Color Maps from matplotlib}.
\newblock R package version 0.5.1.

\bibitem[Harrell, 2020]{R-Hmisc}
Harrell, Jr., F.~E. (2020).
\newblock {\em Hmisc: Harrell Miscellaneous}.
\newblock R package version 4.4-2.

\bibitem[Helmuth and Abdelhady, 2020]{helmuth2020benchmarking}
Helmuth, T. and Abdelhady, A. (2020).
\newblock Benchmarking parent selection for program synthesis by genetic
  programming.
\newblock In {\em Proceedings of the 2020 Genetic and Evolutionary Computation
  Conference Companion}, pages 237--238.

\bibitem[Helmuth and Kelly, 2021]{helmuth_psb2_2021}
Helmuth, T. and Kelly, P. (2021).
\newblock {PSB2}: the second program synthesis benchmark suite.
\newblock In {\em Proceedings of the {Genetic} and {Evolutionary} {Computation}
  {Conference}}, pages 785--794, Lille France. ACM.

\bibitem[Helmuth et~al., 2016a]{helmuth_effects_2016}
Helmuth, T., McPhee, N.~F., and Spector, L. (2016a).
\newblock Effects of {Lexicase} and {Tournament} {Selection} on {Diversity}
  {Recovery} and {Maintenance}.
\newblock In {\em Proceedings of the 2016 on {Genetic} and {Evolutionary}
  {Computation} {Conference} {Companion} - {GECCO} '16 {Companion}}, pages
  983--990, Denver, Colorado, USA. ACM Press.

\bibitem[Helmuth et~al., 2016b]{helmuth_lexicase_2016}
Helmuth, T., McPhee, N.~F., and Spector, L. (2016b).
\newblock Lexicase {Selection} for {Program} {Synthesis}: {A} {Diversity}
  {Analysis}.
\newblock In Riolo, R., Worzel, W., Kotanchek, M., and Kordon, A., editors,
  {\em Genetic {Programming} {Theory} and {Practice} {XIII}}, pages 151--167.
  Springer International Publishing, Cham.
\newblock Series Title: Genetic and Evolutionary Computation.

\bibitem[Helmuth et~al., 2020]{helmuth_importance_2020}
Helmuth, T., Pantridge, E., and Spector, L. (2020).
\newblock On the importance of specialists for lexicase selection.
\newblock {\em Genetic Programming and Evolvable Machines}.

\bibitem[Helmuth and Spector, 2015]{helmuth_general_2015}
Helmuth, T. and Spector, L. (2015).
\newblock General {Program} {Synthesis} {Benchmark} {Suite}.
\newblock In {\em Proceedings of the 2015 on {Genetic} and {Evolutionary}
  {Computation} {Conference} - {GECCO} '15}, pages 1039--1046, Madrid, Spain.
  ACM Press.

\bibitem[Helmuth and Spector, 2020]{helmuth_explaining_2020}
Helmuth, T. and Spector, L. (2020).
\newblock Explaining and {Exploiting} the {Advantages} of {Down}-sampled
  {Lexicase} {Selection}.
\newblock In {\em The 2020 {Conference} on {Artificial} {Life}}, pages
  341--349, Online. MIT Press.

\bibitem[Helmuth and Spector, 2021]{helmuth_problem-solving_2021}
Helmuth, T. and Spector, L. (2021).
\newblock Problem-solving benefits of down-sampled lexicase selection.
\newblock {\em arXiv:2106.06085 [cs]}.
\newblock arXiv: 2106.06085. To be published in Artificial Life Journal.

\bibitem[Helmuth et~al., 2015]{helmuth2015solving}
Helmuth, T., Spector, L., and Matheson, J. (2015).
\newblock Solving uncompromising problems with lexicase selection.
\newblock {\em IEEE Transactions on Evolutionary Computation}, 19(5):630--643.

\bibitem[Hernandez et~al., 2019]{hernandez2019random}
Hernandez, J.~G., Lalejini, A., Dolson, E., and Ofria, C. (2019).
\newblock Random subsampling improves performance in lexicase selection.
\newblock In {\em Proceedings of the Genetic and Evolutionary Computation
  Conference Companion}, pages 2028--2031.

\bibitem[Hernandez et~al., 2021]{supplemental_material}
Hernandez, J.~G., Lalejini, A., and Ofria, C. (2021).
\newblock {Supplemental Material GitHub Repository}.
\newblock doi: 10.5281/zenodo.5020769. url:
  https://doi.org/10.5281/zenodo.5020769.

\bibitem[Hooker, 1995]{hooker_testing_1995}
Hooker, J.~N. (1995).
\newblock Testing heuristics: {We} have it all wrong.
\newblock {\em Journal of Heuristics}, 1(1):33--42.

\bibitem[Jundt and Helmuth, 2019]{jundt_comparing_2019}
Jundt, L. and Helmuth, T. (2019).
\newblock Comparing and combining lexicase selection and novelty search.
\newblock In {\em Proceedings of the {Genetic} and {Evolutionary} {Computation}
  {Conference}}, pages 1047--1055, Prague Czech Republic. ACM.

\bibitem[Kassambara, 2021]{R-rstatix}
Kassambara, A. (2021).
\newblock {\em rstatix: Pipe-Friendly Framework for Basic Statistical Tests}.
\newblock R package version 0.7.0.

\bibitem[La~Cava et~al., 2019]{la_cava_probabilistic_2019}
La~Cava, W., Helmuth, T., Spector, L., and Moore, J.~H. (2019).
\newblock A {Probabilistic} and {Multi}-{Objective} {Analysis} of {Lexicase}
  {Selection} and $\epsilon$-{Lexicase} {Selection}.
\newblock {\em Evolutionary Computation}, 27(3):377--402.

\bibitem[La~Cava et~al., 2016]{la2016epsilon}
La~Cava, W., Spector, L., and Danai, K. (2016).
\newblock Epsilon-lexicase selection for regression.
\newblock In {\em Proceedings of the Genetic and Evolutionary Computation
  Conference 2016}, pages 741--748.

\bibitem[Lalejini and Hernandez, 2021]{supplemental_data}
Lalejini, A.~M. and Hernandez, J.~G. (2021).
\newblock Experiment data.
\newblock {doi: 10.17605/OSF.IO/XPJFT. url: osf.io/xpjft}.

\bibitem[Lehman and Stanley, 2008]{lehman_exploiting_2008}
Lehman, J. and Stanley, K.~O. (2008).
\newblock Exploiting open-endedness to solve problems through the search for
  novelty.
\newblock In {\em Proceedings of the {Eleventh} {International} {Conference} on
  {Artificial} {Life} ({Alife} {XI})}. MIT Press.

\bibitem[Lehman and Stanley, 2011]{lehman_abandoning_2011}
Lehman, J. and Stanley, K.~O. (2011).
\newblock Abandoning {Objectives}: {Evolution} {Through} the {Search} for
  {Novelty} {Alone}.
\newblock {\em Evolutionary Computation}, 19(2):189--223.

\bibitem[Metevier et~al., 2019]{metevier_lexicase_2019}
Metevier, B., Saini, A.~K., and Spector, L. (2019).
\newblock Lexicase {Selection} {Beyond} {Genetic} {Programming}.
\newblock In Banzhaf, W., Spector, L., and Sheneman, L., editors, {\em Genetic
  {Programming} {Theory} and {Practice} {XVI}}, pages 123--136. Springer
  International Publishing, Cham.
\newblock Series Title: Genetic and Evolutionary Computation.

\bibitem[Moore and McKinley, 2016]{moore_comparison_2016}
Moore, J.~M. and McKinley, P.~K. (2016).
\newblock A {Comparison} of {Multiobjective} {Algorithms} in {Evolving}
  {Quadrupedal} {Gaits}.
\newblock In Tuci, E., Giagkos, A., Wilson, M., and Hallam, J., editors, {\em
  From {Animals} to {Animats} 14}, volume 9825, pages 157--169. Springer
  International Publishing, Cham.
\newblock Series Title: Lecture Notes in Computer Science.

\bibitem[Moore and Stanton, 2017]{moore_lexicase_2017}
Moore, J.~M. and Stanton, A. (2017).
\newblock Lexicase selection outperforms previous strategies for incremental
  evolution of virtual creature controllers.
\newblock In {\em Proceedings of the 14th {European} {Conference} on
  {Artificial} {Life} {ECAL} 2017}, pages 290--297, Lyon, France. MIT Press.

\bibitem[Neuwirth, 2014]{R-RColorBrewer}
Neuwirth, E. (2014).
\newblock {\em RColorBrewer: ColorBrewer Palettes}.
\newblock R package version 1.1-2.

\bibitem[Ofria et~al., 2020]{charles_ofria_2020_empirical}
Ofria, C., Moreno, M.~A., Dolson, E., Lalejini, A., Rodriguez-Papa, S., Fenton,
  J., Perry, K., Jorgensen, S., Hoffman, R., Miller, R., Edwards, O.~B.,
  Stredwick, J., G, N.~C., Clemons, R., Vostinar, A., Moreno, R., Schossau, J.,
  Zaman, L., and Rainbow, D. (2020).
\newblock {Empirical: A scientific software library for research, education,
  and public engagement}.
\newblock doi: 10.5281/zenodo.4141943.

\bibitem[Orzechowski et~al., 2018]{orzechowski_where_2018}
Orzechowski, P., La~Cava, W., and Moore, J.~H. (2018).
\newblock Where are we now? a large benchmark study of recent symbolic
  regression methods.
\newblock In {\em Proceedings of the {Genetic} and {Evolutionary} {Computation}
  {Conference}}, pages 1183--1190, Kyoto Japan. ACM.

\bibitem[{R Core Team}, 2020]{r_lang}
{R Core Team} (2020).
\newblock {\em R: A Language and Environment for Statistical Computing}.
\newblock R Foundation for Statistical Computing, Vienna, Austria.

\bibitem[Spector, 2012]{spector_assessment_2012}
Spector, L. (2012).
\newblock Assessment of problem modality by differential performance of
  lexicase selection in genetic programming: a preliminary report.
\newblock In {\em Proceedings of the fourteenth international conference on
  {Genetic} and evolutionary computation conference companion - {GECCO}
  {Companion} '12}, page 401, Philadelphia, Pennsylvania, USA. ACM Press.

\bibitem[Spector et~al., 2018]{la_cava_relaxations_2018}
Spector, L., Cava, W.~L., Shanabrook, S., Helmuth, T., and Pantridge, E.
  (2018).
\newblock Relaxations of {Lexicase} {Parent} {Selection}.
\newblock In Banzhaf, W., Olson, R.~S., Tozier, W., and Riolo, R., editors,
  {\em Genetic {Programming} {Theory} and {Practice} {XV}}, pages 105--120.
  Springer International Publishing, Cham.

\bibitem[Wickham, 2019]{R-tidyverse}
Wickham, H. (2019).
\newblock {\em tidyverse: Easily Install and Load the Tidyverse}.
\newblock R package version 1.3.0.

\bibitem[Wickham et~al., 2021]{R-ggplot2}
Wickham, H., Chang, W., Henry, L., Pedersen, T.~L., Takahashi, K., Wilke, C.,
  Woo, K., Yutani, H., and Dunnington, D. (2021).
\newblock {\em ggplot2: Create Elegant Data Visualisations Using the Grammar of
  Graphics}.
\newblock R package version 3.3.4.

\bibitem[Wilke, 2020]{R-cowplot}
Wilke, C.~O. (2020).
\newblock {\em cowplot: Streamlined Plot Theme and Plot Annotations for
  ggplot2}.
\newblock R package version 1.1.0.

\bibitem[Xie, 2020a]{R-bookdown}
Xie, Y. (2020a).
\newblock {\em bookdown: Authoring Books and Technical Documents with R
  Markdown}.
\newblock R package version 0.21.

\bibitem[Xie, 2020b]{R-knitr}
Xie, Y. (2020b).
\newblock {\em knitr: A General-Purpose Package for Dynamic Report Generation
  in R}.
\newblock R package version 1.30.

\bibitem[Zhu, 2021]{R-kableExtra}
Zhu, H. (2021).
\newblock {\em kableExtra: Construct Complex Table with kable and Pipe Syntax}.
\newblock R package version 1.3.4.

\end{thebibliography}
